%% file: main.tex
\documentclass[]{assets/template}
\usepackage[utf8]{inputenc}
\usepackage{setspace}
\usepackage[dvipsnames]{xcolor} % 只加载一次
\usepackage{amssymb, amsmath, amsthm, mathrsfs, mathtools, bbm, dsfont, mathabx}

\usepackage{float}
\usepackage{ragged2e}
\usepackage{array}
\usepackage[section]{placeins}
\usepackage{graphicx}
\usepackage{wrapfig}
\usepackage{caption}
\usepackage{subcaption}
\usepackage{booktabs}
\usepackage{threeparttable}
\usepackage{multirow}
\usepackage{makecell}
\usepackage{tabularx}
\usepackage{colortbl}
\usepackage{arydshln}
\usepackage{float}
\usepackage{adjustbox}

\usepackage{algorithm}
\usepackage{algpseudocode} % 这会加载 algorithmicx

\usepackage{enumitem}
\usepackage{markdown}
\usepackage{pifont}
\usepackage{fontawesome5}
\usepackage{listings}
\usepackage{siunitx}

\PassOptionsToPackage{most,breakable}{tcolorbox}
\tcbuselibrary{listings,skins,breakable}
\tcbset{listing engine=listings}

\usepackage{tikz}
\usetikzlibrary{fit, calc}

\usepackage{natbib}
\usepackage{hyperref}
\usepackage{url}
\usepackage{cleveref}

\definecolor{lightgray}{gray}{0.9}

\usepackage{bxcoloremoji}

\addtocontents{toc}{\protect\setcounter{tocdepth}{0}}

\newcommand{\hc}[1]{green!\fpeval{((#1-50)/50)^2*100}!white!0}
\newcommand{\heat}[1]{\cellcolor{\hc{#1}}#1}
\newcommand{\heatul}[1]{\cellcolor{\hc{#1}}\underline{#1}}
\newcommand{\heatbf}[1]{\cellcolor{\hc{#1}}\textbf{#1}}

\definecolor{modelbg}{rgb}{0.9, 0.95, 1.0}
\definecolor{metagreen}{HTML}{2E8B57} % SeaGreen
\definecolor{lightblue}{RGB}{210, 220, 250}
\definecolor{midblue}{RGB}{86, 108, 184}
\definecolor{msblue}{RGB}{0,102,204}
\definecolor{GREEN}{RGB}{69,138,0}
\definecolor{RED}{RGB}{200,0,0}
\definecolor{mypink}{HTML}{FF3366}

\newcommand{\ours}{RPG-Encoder}
\addtocontents{toc}{\protect\setcounter{tocdepth}{0}}
\newdateformat{usvardate}{\monthname[\THEMONTH] \space \THEDAY, \space \THEYEAR}

\title{Closing the Loop: Universal Repository Representation with RPG-Encoder}

\author{
Jane Luo$^{1, \ddagger,\,*}$,
Chengyu yin$^{1\,\ddagger,*}$, 
Xin Zhang$^{1\, *,\dagger}$, 
Qingtao Li$^{1}$, 
Steven Liu$^{1,\ddagger}$,
Yiming Huang$^{2}$,\\
\textbf{Jie Wu}$^{3,\ddagger}$, 
\textbf{Hao Liu}$^{1,\ddagger}$, 
\textbf{Yangyu Huang}$^{1}$, 
\textbf{Yu Kang}$^{1}$,
\textbf{Fangkai Yang}$^{1}$, 
\textbf{Ying Xin}$^{1}$,
\textbf{Scarlett Li}$^{1}$, \\ 
$^1$Microsoft Research Asia \quad $^2$UCSD \quad $^3$Tsinghua University \\
$^{*}$ Equal contribution ~$^{\dagger}$ Corresponding author
$^{\ddagger}$ Work done during internships at Microsoft
}

\begin{document}
\input{sections/0_abstract}

\maketitle
\input{sections/1_introduction}
\input{sections/related_work}
\input{sections/2_method}
\input{sections/3_experiments}
\input{sections/4_ablation}
\input{sections/4_analysis}

\input{sections/5_conclusion}
\bibliographystyle{assets/plainnat}
\bibliography{reference}

\clearpage
\input{sections/appendix}
\end{document}

%% file: sections/0_abstract.tex
\begin{abstract}
\label{sec:abstract}

Current repository agents encounter a reasoning disconnect due to fragmented representations, as existing methods rely on isolated API documentation or dependency graphs that lack semantic depth. We consider repository comprehension and generation to be inverse processes within a unified cycle: generation expands intent into implementation, while comprehension compresses implementation back into intent. To address this, we propose \ours{}, a framework that generalizes the Repository Planning Graph (RPG) from a static generative blueprint into a unified, high-fidelity representation. \ours{} closes the reasoning loop through three mechanisms: (1) Encoding raw code into the RPG that combines lifted semantic features with code dependencies; (2) Evolving the topology incrementally to decouple maintenance costs from repository scale, reducing overhead by 95.7\%; and (3) Operating as a unified interface for structure-aware navigation. In evaluations, \ours{} establishes state-of-the-art localization performance on SWE-bench Verified with 93.7\% Acc@5 and exceeds the best baseline by over 10\% in localization accuracy on SWE-bench Live Lite. These results highlight our superior fine-grained precision in complex codebases. Furthermore, it achieves 98.5\% reconstruction coverage on RepoCraft, confirming RPG's high-fidelity capacity to mirror the original codebase and closing the loop between intent and implementation.

\vspace{5pt}
\coloremojicode{1F4C5}~ \textbf{Date}: \usvardate\today

\faGithub~ \textbf{Code}: \href{https://github.com/microsoft/RPG-ZeroRepo}{https://github.com/microsoft/RPG-ZeroRepo}

\coloremojicode{1F310}~ \textbf{Project}: \href{https://ayanami2003.github.io/RPG-Encoder/}{https://ayanami2003.github.io/RPG-Encoder/}

\coloremojicode{1F4C4}~ \textbf{Prior Work}: \href{https://arxiv.org/abs/2509.16198}{https://arxiv.org/abs/2509.16198}

\coloremojicode{2709}~ \textbf{Correspondence}: \href{janeluo1210@163.com}{janeluo1210@163.com}; \href{xinzhang3@microsoft.com}{xinzhang3@microsoft.com}

\end{abstract}

%% file: sections/1_introduction.tex
\section{Introduction}
\label{sec:intro}

\begin{figure}[ht] 
\centering
\includegraphics[width=\linewidth]{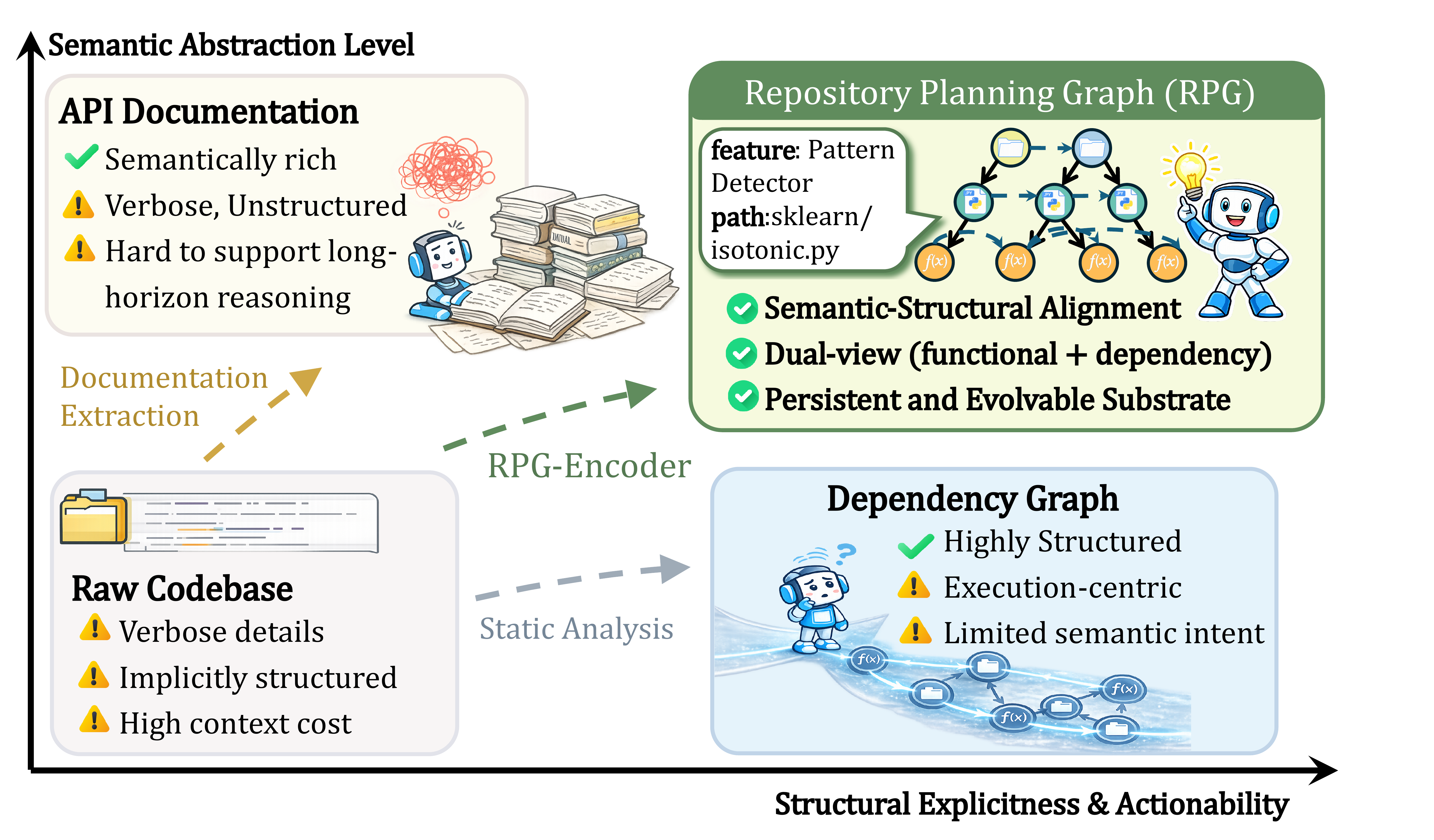}
\caption{Comparison of code representations regarding semantic abstraction and structural explicitness. Unlike approaches limited to a single dimension, RPG achieves dual-view alignment, combining semantic richness with structural actionability.} 
\label{fig:intention}
\vspace*{-5pt} 
\end{figure}

Repository-level software engineering relies on agents navigating complex dependencies and reasoning about high-level architectural intent~\citep{wang2025improving, zhao2025towards}. However, as illustrated in Figure~\ref{fig:intention}, existing approaches suffer from a reasoning gap due to fragmented representations: \textbf{API Documentation} focuses on semantic intent~\citep{luo2024repoagent, chen2025llms} but lacks global navigability, forcing models to infer architectural connectivity~\citep{chen2025towards, jain2025mitigating}. Conversely, \textbf{Dependency Graph} captures structural logic~\citep{ouyang2024repograph, ma2024understand} but provide limited semantic information~\citep{borowski2024semantic, cheng2024semantic}, leaving agents to follow execution paths without reflecting the underlying rationale~\citep{jiang2025cosil}. Furthermore, maintaining consistency incurs prohibitive overhead: documentation is prone to semantic drift~\citep{tan2024detecting}, while static graphs capture syntactic updates but often overlook logical implications~\citep{groninger2025changeguard}.

We observe that this reasoning disconnect is not merely a failure of individual tools, but a systemic consequence of treating repository understanding as an isolated, unidirectional task. Fundamentally, this occurs because current approaches ignore the inherent symmetry of software engineering.
We argue that repository comprehension and generation constitute inverse pathways within a unified reasoning cycle: generation expands sparse intent into detailed code, whereas comprehension must compress noisy implementation back into high-level intent. Consequently, bridging this gap requires a \textbf{unified Intermediate Representation} that fuses the semantic density of documentation with the topological rigor of dependency graphs. The Repository Planning Graph (RPG)~\citep{luo2025rpg_zerorepo} emerges as a suitable representation for this unification. Having served as a generative blueprint for intent-to-code, it possesses the dual-view structure needed for the inverse code-to-intent journey. This motivates our fundamental inquiry: \textit{Can the RPG be generalized to serve as a unified, high-fidelity representation for existing repositories, thereby closing the loop?}

To realize this vision, we propose \ours{}, a framework that transforms the RPG from a static generative blueprint into a dynamic, bidirectional representation. We implement this through three cohesive mechanisms:
(1) \textbf{Encoding:} We introduce a semantic lifting protocol that projects code into the RPG. Nodes combine functional descriptions with code metadata, while edges encode hierarchy and static dependencies, yielding an interpretable and verifiable representation.
(2) \textbf{Evolution:} We design an incremental mechanism that parses commit diffs to update the RPG. This keeps semantics synchronized with implementation without re-generation.
(3) \textbf{Operation:} We establish the RPG as a unified interface for structure-aware reasoning. It serves as a topological map, enabling traversal between high-level intent and low-level execution logic.

To evaluate the extracted RPG, we conduct a dual-task evaluation on two critical dimensions: navigational utility and representational fidelity. (1) In \textbf{Repository Understanding}, \ours{} with Claude-4.5-Sonnet~\citep{anthropic_claude_sonnet_4_5} demonstrates superior function-level localization performance, achieving 93.7\% Acc@5 on SWE-bench Verified~\citep{jimenez2023swe} and exceeding the best baseline by over 10\% in localization accuracy on SWE-bench Live Lite~\citep{zhang2025swe}. This confirms that coupling semantic features with topology significantly strengthens fine-grained localization. (2) In \textbf{Repository Reconstruction}, \ours{} outperforms API documentation by providing an explicitly ordered blueprint. Guided by topological constraints, \ours{} reconstructs repositories with 98.5\% coverage (+24.3\% over baselines) and 86.0\% pass rate on RepoCraft~\citep{luo2025rpg_zerorepo}. In contrast, documentation lacks structural guidance and recovers only $\sim 17\%$ of the original code volume, proving that RPG serves as a structured representation that effectively preserves complete repository semantics. Analysis confirms that semantic features are essential for effective exploration, and our incremental strategy reduces maintenance costs by 95.7\% without incurring semantic drift.

Our contributions are summarized as follows:
\begin{itemize}
    \item We generalize the Repository Planning Graph (RPG) into a unified representation that closes the loop between comprehension and generation, theoretically grounding repository reasoning as a unified reasoning cycle where semantic intent and structural dependencies are bidirectionally linked. 

    \item We introduce \ours{}, a framework that implements a semantic lifting protocol to recover high-level intent from code and supports sustainable evolution via differential updates, decoupling maintenance costs from repository scale.

    \item We validate \ours{} on dual tasks: establishing SOTA performance in repository understanding to demonstrate superior navigational utility, and achieving 98.5\% coverage in repository reconstruction to verify its high-fidelity representational capacity.
\end{itemize}

%% file: sections/related_work.tex
\section{Related Work}
\label{sec:related}

\textbf{Repository Generation.} Research has transitioned from localized file completion~\citep{wang2025epicoder, li2023starcoder} to systemic workflows that emphasize architectural coherence. Multi-agent frameworks like MetaGPT~\citep{hong2024metagpt} and paper-to-code systems~\citep{seo2025paper2code, lin2025autop2c} utilize role-based abstraction to manage complexity. More recently, Commit0~\citep{zhao2024commit0} introduces a library-level reconstruction paradigm starting from near-zero implementations, while terminal-based agents such as Claude Code~\citep{anthropic2025claude_code} and Gemini CLI~\citep{google2025gemini_cli} facilitate iterative "build-test-fix" cycles in real-world environments. RPG~\citep{luo2025rpg_zerorepo} advances this by utilizing structured planning graphs to ground generative intent in execution dependencies, ensuring that synthesized repositories remain topologically valid.

\textbf{Repository Understanding.} Current paradigms shift from passive retrieval to active, structure-aware exploration. Early iterative methods~\citep{zhang2023repocoder, xia2024agentless} have been augmented by graph-guided navigation frameworks~\citep{ouyang2024repograph, liu2025codexgraph}. To address high-density codebases, LocAgent~\citep{chen2025locagent} and KGCompass~\citep{yang2025kgcompass} leverage explicit dependency schemas and knowledge graphs to prune the search space. Furthermore, RepoHyper~\citep{phan2025repohyper} explores long-context summarization for global semantic grasp, while agents like OrcaLoca~\citep{yu2025orcaloca} and CoSIL~\citep{jiang2025cosil} integrate dynamic execution signals for precise fault localization. These systems increasingly rely on "browse-edit-run" loops~\citep{yang2024swe} and are rigorously benchmarked on real-world issue-solving datasets like SWE-bench Verified~\citep{jimenez2023swe}. Despite these advancements, existing methods often suffer from fragmented representations: dependency graphs lack semantic depth, while semantic retrievers lack topological precision. \ours{} bridges this gap by coupling dense semantic features with structural constraints, enabling fine-grained localization that is both intent-driven and execution-grounded.

%% file: sections/2_method.tex
\section{Method}
\label{sec:method}
\begin{figure}[H]
    \centering
    \includegraphics[width=\linewidth]{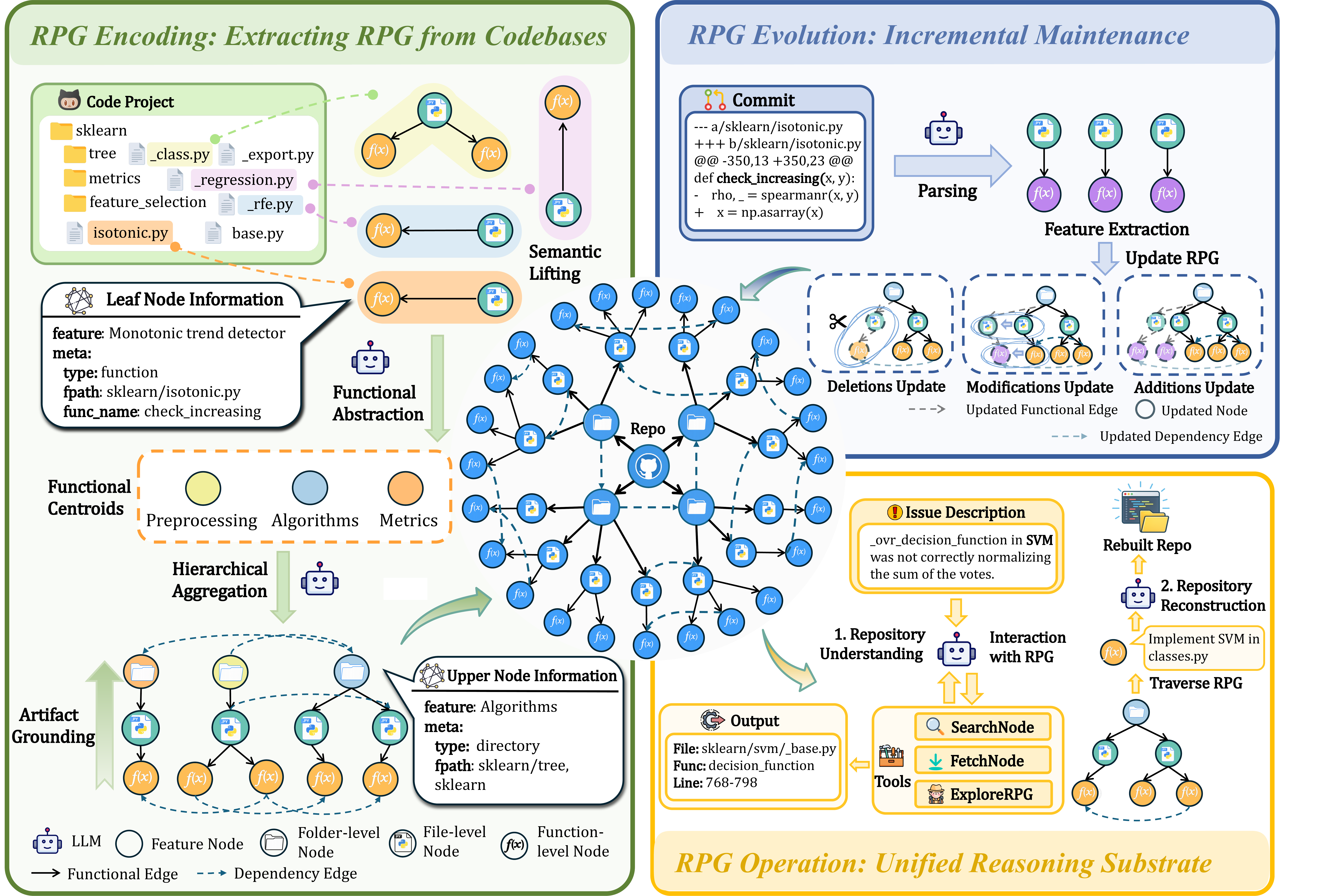}
    \caption{Overview of the RPG-Encoder. The pipeline bridges Code and RPG via three stages: \textbf{Encoding} lifts code into a semantic topology; \textbf{Evolution} handles incremental updates via commits; and \textbf{Operation} provides a unified interface for agentic reasoning.\protect\footnotemark}
    \label{fig:pipeline}
\end{figure}

To establish RPG as a unified and high-fidelity Intermediate Representation, we introduce the \ours{}. By mapping implementation back to semantic space (Code $\to$ RPG), it completes the representation loop. As illustrated in Figure~\ref{fig:pipeline}, our methodology comprises: (1) Encoding for RPG extraction; (2) Evolution for incremental maintenance; and (3) Operation as a unified reasoning substrate.

\subsection{RPG Encoding: Extracting RPG from Codebases}
To transform a raw codebase into an actionable substrate, we model extraction as a pipeline that converts implementation details into a compact, structured semantic index for high-level reasoning. This process reconstructs the system topology in three phases. More details are in Appendix~\ref{app:extraction}. 

\paragraph{RPG Structure} Refining prior definitions \citep{luo2025rpg_zerorepo}, we define RPG as a hierarchical, dual-view graph $\mathcal{G} = (\mathcal{V}, \mathcal{E})$. The node set $\mathcal{V} = \mathcal{V}_{H} \cup \mathcal{V}_{L}$ distinguishes High-level Nodes representing architectural directories from Low-level Nodes comprising atomic implementations such as files, classes, and functions. Each node $v = (f, \mathbf{m}) \in \mathcal{V}$ pairs a semantic feature $f$ describing functionality (e.g., \textit{handles authentication}) with structural metadata $\mathbf{m}$ encoding code entity attributes like type and path. The edge set $\mathcal{E}$ integrates two perspectives: (1) Functional edges $\mathcal{E}_{\text{feature}}$ establishing teleological hierarchy; and (2) Dependency edges $\mathcal{E}_{\text{dep}}$ mapping logical interactions including imports and calls. This duality enables the agent to perceive the repository as both a functional and executable network.

\footnotetext{The Python icon is the official Python logo; see \url{https://www.python.org/community/logos/}.}

\vspace*{-15pt}
\paragraph{Phase 1: Semantic Lifting}
To bridge the granularity mismatch between verbose implementation and functional intent, the extraction process first lifts the codebase into a discrete registry of Low-level Nodes ($\mathcal{V}_L$). For each file, the system extracts semantic features $f$ for individual functions and classes, mapping them to behavioral signatures while retaining their code-level attributes as metadata $\mathbf{m}$. Subsequently, these fine-grained features are synthesized into a holistic summary representing the file's overall functionality. This summarization process naturally establishes functional edges $\mathcal{E}_{\text{feature}}$ between the file-level node and its constituent function-level node. This phase concludes by producing a semantically grounded implementation index, serving as a robust representation for higher-level reasoning.
 
\paragraph{Phase 2: Semantic Structure Reorganization}
Physical folder-file organization is often dictated by technical constraints rather than functional boundaries, inducing structural entanglement. To mitigate this, we construct the High-level Node set $\mathcal{V}_H$ by recovering the latent functional topology from implementation units ($\mathcal{V}_L$). (1) \textbf{Functional Abstraction:} To ensure the global repository state fits within the LLM context window, we perform granularity-based input compression. Instead of raw implementation, the LLM only consumes concise semantic features $f$ of file-level nodes, excluding function-level details. This condensed view allows the model to analyze the complete repository-wide semantic manifold to induce abstract functional centroids (e.g., \textit{Data Preprocessing}) that define the root pillars of the hierarchy. (2) \textbf{Hierarchical Aggregation:} We recursively link nodes from $\mathcal{V}_L$ to these centroids. To ensure structural stability, each node’s placement is determined by a semantic compatibility check: the LLM evaluates the fit between a node’s $f$ and the centroid’s definition, instantiating intermediate nodes (e.g., routing \textit{StandardScaler} via \textit{Normalization} to \textit{Preprocessing}) to bridge the hierarchy when a direct link lacks granularity. Together, these nodes constitute the High-level Node set $\mathcal{V}_H$, establishing explicit parent-child functional edges. This yields a complete functional graph where each $v \in \mathcal{V}_H$ possesses semantic feature $f$ but lacks structural metadata $\mathbf{m}$ required to link it to physical code entities.

\paragraph{Phase 3: Artifact Grounding}
To transform the abstract hierarchy into a substrate, this phase anchors the functional manifold to physical artifacts and execution logic. We first populate the missing metadata $\mathbf{m}$ for nodes in $\mathcal{V}_H$ through bottom-up propagation, utilizing a Lowest Common Ancestor (LCA) mechanism (detailed in Appendix \ref{app:extraction_grounding}) to compute the minimal directory scope shared by each cluster's descendants. This mapping ensures that abstract features such as \textit{Data Preprocessing} are tied to code paths like \texttt{sklearn/preprocessing}. Subsequently, to transition from a semantic hierarchy to an implementation map, we inject dependency edges $\mathcal{E}_{\text{dep}}$ (e.g., imports, calls) via AST analysis. This integration completes the RPG, yielding a unified representation that enables traceability between high-level functional intent and the executable code.

\subsection{RPG Evolution: Incremental Maintenance}
To reduce the cost of full re-generation, we maintain $\mathcal{G}$ incrementally and reserve global reconstruction for major refactoring. For routine updates, we perform online graph editing to keep the RPG synchronized, as illustrated in Figure~\ref{fig:pipeline} (top-right) and detailed in Appendix~\ref{app:evolution}.

\paragraph{Commit-Level Feature Extraction}
We parse raw commit data to extract semantic features strictly for affected code fragments, avoiding full reprocessing. This yields a set of discrete Feature Nodes representing the delta state, which serves as the direct input for graph operations.

\paragraph{RPG Updates}
Based on the diff type, we execute three atomic update protocols to maintain the RPG structure:
(1) \textbf{Deletions}: We remove nodes for deleted files or functions and recursively prune empty parent categories in $\mathcal{V}H$ to maintain hierarchical integrity.
(2) \textbf{Modifications}: We re-generate the semantic description $f$ for modified entities. To avoid structural instability, a node’s position is updated only if the LLM detects a functional intent shift that violates its parent’s semantic scope (e.g., a utility function evolving into a core algorithm). This check serves as a semantic threshold to prevent minor implementation changes from triggering costly structural migrations.
(3) \textbf{Additions}: We create nodes for new entities and insert them into the hierarchy by matching their semantics against existing functional centroids. Finally, we perform a localized dependency update, re-parsing affected ASTs to refresh $\mathcal{E}{\text{dep}}$ and align connectivity with the execution flow.

\subsection{RPG Operation: Unified Reasoning Substrate}
We deploy RPG as a Unified Representation providing a queryable index of the codebase. Structurally, it functions as a heterogeneous graph where Functional and Dependency Views are partitioned by edge types ($\mathcal{E}_{\text{feature}}$ and $\mathcal{E}_{\text{dep}}$) but share a unified node set, enabling seamless context switching during retrieval. More details are in Appendix \ref{app:operation}
\paragraph{Unified Agentic Tool}
We define three core tools to operate on the RPG's nodes and edges:
\begin{itemize}
    
    \item \textbf{SearchNode}: Performs global node-level retrieval by matching intent against semantic features $f$ or filtering metadata $\mathbf{m}$, allowing the agent to precisely localize entry points across both views.
    
    \item \textbf{FetchNode}: Executes node-level data retrieval. Given $v$, it extracts the attribute tuple $(f, \mathbf{m})$ and raw source code to provide the ground truth for inspection.
    
    \item \textbf{ExploreRPG}: Facilitates cross-view traversal along edges $\mathcal{E}$. While $\mathcal{E}_{\text{dep}}$ is strictly constructed via static AST analysis, its integration with the semantic hierarchy in $\mathcal{V}_H$ provides a robust topological skeleton that guides the agent through complex execution flows without the noise of unstructured search.
\end{itemize}

This toolset enables multi-dimensional navigation by integrating functional intent with physical implementation, facilitating precise context discovery through semantic and dependency structures.

\paragraph{Efficient Structured Representation}
RPG reduces information overload by representing the repository as a substrate with two roles: (1) Knowledge Source: RPG stores feature descriptions and metadata for each node, capturing \textit{what} the code does without parsing implementations. (2) Process Encoder: RPG induces a topological order via functional edges ($\mathcal{E}_{\text{feature}}$) and dependency edges ($\mathcal{E}_{\text{dep}}$), exposing causality and hierarchy essential for architectural comprehension.

%% file: sections/3_experiments.tex
\vspace*{-5pt}
\section{Experiments Setup}
\label{sec:exp_setup}
We evaluate RPG on two tasks to assess its semantic grasp and structural completeness: (1) Repository Understanding, testing navigation and localization capabilities; and (2) Repository Reconstruction, verifying the fidelity and losslessness of the encoded information.
\vspace*{-5pt}
\subsection{Repository Understanding}
We assess RPG as a navigational substrate through rigorous localization tasks. More details are in Appendix~\ref{app:under_detail}. 
\vspace*{-20pt}
\paragraph{Benchmark.}
We evaluate on two benchmarks: SWE-bench Verified~\citep{jimenez2023swe}, a human-validated subset ensuring solvability with 500 examples from 12 repositories; and SWE-bench-Live Lite~\citep{zhang2025swe}, mitigating contamination using recent issues, comprising 300 examples across 70 repositories.
\vspace*{-5pt}
\paragraph{Baselines.}We compare against baselines leveraging diverse structural priors: Agentless~\citep{xia2024agentless} operates via hierarchical text-based narrowing without graph priors; LocAgent~\citep{chen2025locagent} leverages explicit dependency graphs for guided traversal; CoSIL~\citep{jiang2025cosil} performs iterative search over static code structures; and OrcaLoca~\citep{yu2025orcaloca} integrates dynamic execution signals with agentic planning.
\vspace*{-5pt}
\paragraph{Evaluation Metrics.}
We adopt standard metrics: Acc@k ($k\!\in\!\{1,5\}$) checks if a ground-truth target is in top-$k$ predictions~\citep{jiang2025cosil}; and Precision/Recall quantify overlap. Given predicted set $P$ and ground-truth $G$, we define $\mathrm{Precision}=|P\cap G|/|P|$ and $\mathrm{Recall}=|P\cap G|/|G|$.
\vspace*{-5pt}
\paragraph{Implementation Details.}
We use GPT-4o~\citep{gpt4o-system-card} to parse and incrementally update the RPG. Backbone models include o3-mini~\citep{o3-mini}, GPT-4o~\citep{gpt4o-system-card}, GPT-4.1~\citep{gpt4.1}, GPT-5~\citep{gpt5-system-card}, DeepSeek-V3.1~\citep{deepseek-v3.1}, and Claude-4.5-sonnet~\citep{anthropic_claude_sonnet_4_5}. \ours{} operates with a 40-step limit. Baselines follow configurations (detailed in Appendix~\ref{app:under_params_baselines}). All runs are averaged over 3 times.
\vspace*{-5pt}
\subsection{Repository Reconstruction}
We use reconstruction to verify lossless, topologically ordered RPG information. Details are in Appendix~\ref{app:recon_detail}.
\vspace*{-5pt}
\paragraph{Benchmark.} We adapt RepoCraft~\citep{luo2025rpg_zerorepo} for controlled reconstruction, aiming to rebuild target repositories (e.g., \texttt{Requests}) with ground-truth functionality. To isolate representational fidelity, we compare Official API Documentation with RPG. We focus on representation sources rather than search-based agents (e.g., LocAgent), since reconstruction requires a comprehensive blueprint instead of iterative localization.
\vspace*{-5pt}
\paragraph{Baselines.} We configure ZeroRepo~\citep{luo2025rpg_zerorepo} in two modes: (1) \textbf{ZeroRepo-Doc (Baseline):} The agent references API documentation, autonomously managing progress and objectives via Test-Driven Development. (2) \textbf{ZeroRepo-RPG (Ours):} We utilize the extracted RPG for direct repository generation, where it serves as the exclusive knowledge source and scheduler. Nodes are processed in topological order, batching semantically similar nodes to accelerate inference. More details are in Appendix~\ref{app:recon_baseline}.
\vspace*{-5pt}
\paragraph{Evaluation Metrics.} Following RepoCraft, we report: (1) Coverage, the proportion of implemented functional categories; (2) Accuracy (Pass / Vote), unit-test pass accuracy and vote-based check accuracy; and (3) Code Statistics (\#Files, nLOC, Code Tokens) to measure structural similarity and recovered code volume.
\vspace*{-5pt}
\paragraph{Implementation Details.}
We employ GPT-4o~\citep{gpt4o-system-card} for RPG extraction and evaluate reconstruction using GPT-5-mini~\citep{gpt5mini} and GPT-4.1~\citep{gpt4.1}. Following RepoCraft, we also use o3-mini~\citep{o3-mini} for automated evaluation. ZeroRepo-Doc runs without a hard turn limit and stops when the agent judges the documentation to be fully implemented. ZeroRepo-RPG is bounded by the graph and terminates once all RPG-derived nodes are executed. More details are in Appendix~\ref{app:recon_impl}.
\input{tables/loc}

\vspace*{-30pt}
\section{Main Result}
\label{sec:result}
\paragraph{RPG Enhances Fine-Grained Repository Understanding.} Table~\ref{tab:loc_result} demonstrates that RPG consistently improves file-level and function-level localization. On SWE-bench Verified, \ours{} with Claude-4.5 achieves 93.7\% Acc@5 on function level, surpassing the best baseline (OrcaLoca) by 14.4 points, while simultaneously improving Precision by 6.9\% and Recall by 10.7\%. Furthermore, on SWE-bench Live, \ours{} with GPT-5 elevates performance to 87.8\% Acc@5 on function level, outperforming CoSIL by 11.6 points. These results confirm that coupling semantic features with topological constraints enables agents to map high-level intent to specific implementation units. Crucially, this dual-view structure filters irrelevant noise while ensuring comprehensive coverage of target functionalities.

\input{tables/reconstruction}

% \vspace*{-5pt}
\paragraph{RPG Functioning as a Complete Representational Substrate.}
Table~\ref{tab:reconstruction_results} demonstrates RPG's superior fidelity in reproducing complex repository structures. With GPT-5-mini, \ours{} attains 98.5\% Coverage and an 86.0\% Pass Rate, exceeding the documentation-based baseline by over 33 points. Regarding code scale, the baseline generates severely fragmented outputs, capturing only $\sim$17\% of the original volume due to a lack of structural guidance. In contrast, \ours{} reconstructs 550k tokens, a scale comparable to the gold project written by human. This high fidelity proves that RPG serves as a sufficient substrate to ground architectural intent within a valid structural topology, guiding the agent to expand the blueprint into concrete implementation unlike linear API documentation.

%% file: tables/loc.tex
\begin{table*}[h]
\centering
% \scriptsize
\renewcommand{\arraystretch}{1.2}
\caption{
Comprehensive localization results on SWE-bench Verified and SWE-bench Live Lite across File and Function levels.
Acc@k: Accuracy@k. Pre/Rec: Precision/Recall.
\textbf{Bold} indicates the best result, and \underline{Underline} indicates the second best.
}
\label{tab:loc_result}

\resizebox{\textwidth}{!}{%
\begin{tabular}{@{}l cccc cccc | cccc cccc@{}}
\toprule
% === Header Row 1 ===
\multirow{3}{*}{\textbf{Method}} & 
\multicolumn{8}{c}{\textbf{SWE-bench Verified}} & 
\multicolumn{8}{c}{\textbf{SWE-bench Live}} \\
\cmidrule(lr){2-9} \cmidrule(l){10-17}

% === Header Row 2: Granularity===
 & \multicolumn{4}{c}{\textbf{File-level}} & \multicolumn{4}{c}{\textbf{Function-level}} 
 & \multicolumn{4}{c}{\textbf{File-level}} & \multicolumn{4}{c}{\textbf{Function-level}} \\
\cmidrule(lr){2-5} \cmidrule(lr){6-9} 
\cmidrule(lr){10-13} \cmidrule(l){14-17} 

% === Header Row 3: Metrics===
 & Acc@1 & Acc@5 & Pre & Rec & Acc@1 & Acc@5 & Pre & Rec 
 & Acc@1 & Acc@5 & Pre & Rec & Acc@1 & Acc@5 & Pre & Rec \\
\midrule

% ================= Model: o3-mini =================
\rowcolor{modelbg}
\multicolumn{17}{c}{\textbf{Model: o3-mini}} \\

Agentless & \heat{67.1} & \heatul{88.1} & \heat{67.0} & \heatul{64.7} & \heat{34.7} & \heat{60.3} & \heat{39.4} & \heat{33.2} & \heat{54.2} & \heat{78.5} & \heat{55.6} & \heat{47.7} & \heat{28.8} & \heat{54.2} & \heat{39.3} & \heat{25.6} \\

OrcaLoca & \heatul{67.5} & \heat{71.9} & \heatul{68.3} & \heat{64.0} & \heat{46.3} & \heat{52.9} & \heat{48.3} & \heat{41.5} & \heat{35.4} & \heat{38.0} & \heat{36.2} & \heat{27.6} & \heat{23.1} & \heat{26.1} & \heat{25.3} & \heat{15.6} \\

LocAgent & \heat{62.8} & \heat{77.2} & \heat{64.7} & \heat{61.4} & \heat{32.1} & \heat{40.5} & \heat{33.9} & \heat{28.9} & \heat{47.6} & \heat{59.4} & \heat{49.7} & \heat{41.2} & \heat{23.8} & \heat{31.0} & \heat{26.6} & \heat{17.7} \\

CoSIL & \heat{66.5} & \heat{85.7} & \heat{66.2} & \heat{63.6} & \heatul{52.2} & \heatul{73.3} & \heatul{54.7} & \heatul{47.1} & \heatul{60.9} & \heatul{80.8} & \heatul{66.1} & \heatul{54.8} & \heatul{43.8} & \heatul{65.1} & \heatul{51.4} & \heatul{35.6} \\

\textbf{Repo-Enc} & \heatbf{78.3} & \heatbf{91.2} & \heatbf{80.7} & \heatbf{76.8} & \heatbf{58.5} & \heatbf{77.8} & \heatbf{62.9} & \heatbf{55.1} & \heatbf{73.7} & \heatbf{88.2} & \heatbf{77.5} & \heatbf{64.5} & \heatbf{56.5} & \heatbf{75.6} & \heatbf{64.7} & \heatbf{46.9} \\

\hline
\textbf{$\Delta_{\text{best}}$} & 
\color[HTML]{2D885D}\textbf{+10.8} & 
\color[HTML]{2D885D}\textbf{+3.1} & 
\color[HTML]{2D885D}\textbf{+12.4} & 
\color[HTML]{2D885D}\textbf{+12.1} & 
\color[HTML]{2D885D}\textbf{+6.3} & 
\color[HTML]{2D885D}\textbf{+4.5} & 
\color[HTML]{2D885D}\textbf{+8.2} & 
\color[HTML]{2D885D}\textbf{+8.0} & 
\color[HTML]{2D885D}\textbf{+12.8} & 
\color[HTML]{2D885D}\textbf{+7.4} & 
\color[HTML]{2D885D}\textbf{+11.4} & 
\color[HTML]{2D885D}\textbf{+9.7} & 
\color[HTML]{2D885D}\textbf{+12.7} & 
\color[HTML]{2D885D}\textbf{+10.5} & 
\color[HTML]{2D885D}\textbf{+13.3} & 
\color[HTML]{2D885D}\textbf{+11.3} \\

% ================= Model: GPT-4o =================
\rowcolor{modelbg}
\multicolumn{17}{c}{\textbf{Model: GPT-4o}} \\

Agentless & \heat{63.0} & \heat{86.1} & \heat{63.1} & \heat{61.1} & \heat{31.4} & \heat{58.8} & \heat{34.7} & \heat{29.3} & \heat{56.1} & \heat{78.8} & \heat{57.1} & \heat{48.3} & \heat{30.6} & \heat{57.4} & \heat{41.4} & \heat{26.4} \\

OrcaLoca & \heat{64.3} & \heat{69.3} & \heat{65.0} & \heat{61.4} & \heat{39.8} & \heat{53.3} & \heat{42.5} & \heat{36.7} & \heat{42.5} & \heat{47.6} & \heat{45.0} & \heat{34.0} & \heat{28.2} & \heat{37.0} & \heat{32.5} & \heat{21.1} \\

LocAgent & \heatul{71.9} & \heatul{87.9} & \heatul{73.4} & \heatul{69.3} & \heat{40.1} & \heatul{67.4} & \heat{44.8} & \heat{38.1} & \heatul{62.5} & \heatul{80.0} & \heatul{66.8} & \heatul{54.2} & \heat{35.7} & \heat{56.4} & \heat{44.5} & \heatul{29.9} \\

CoSIL & \heat{64.9} & \heat{84.4} & \heat{65.0} & \heat{62.2} & \heatul{43.2} & \heat{66.2} & \heatul{48.2} & \heatul{40.1} & \heat{60.1} & \heat{77.0} & \heat{63.7} & \heat{50.7} & \heatul{41.2} & \heatul{61.6} & \heatul{49.1} & \heat{29.4} \\

\textbf{Repo-Enc} & \heatbf{74.5} & \heatbf{89.6} & \heatbf{77.0} & \heatbf{72.7} & \heatbf{53.1} & \heatbf{76.7} & \heatbf{57.9} & \heatbf{49.5} & \heatbf{69.2} & \heatbf{83.5} & \heatbf{73.2} & \heatbf{60.3} & \heatbf{50.5} & \heatbf{69.4} & \heatbf{59.4} & \heatbf{41.8} \\

\hline
\textbf{$\Delta_{\text{best}}$} & 
\color[HTML]{2D885D}\textbf{+2.6} &
\color[HTML]{2D885D}\textbf{+1.7} &
\color[HTML]{2D885D}\textbf{+3.6} &
\color[HTML]{2D885D}\textbf{+3.4} &
\color[HTML]{2D885D}\textbf{+9.9} &
\color[HTML]{2D885D}\textbf{+9.3} &
\color[HTML]{2D885D}\textbf{+9.7} &
\color[HTML]{2D885D}\textbf{+9.4} &
\color[HTML]{2D885D}\textbf{+6.7} &
\color[HTML]{2D885D}\textbf{+3.5} &
\color[HTML]{2D885D}\textbf{+6.4} &
\color[HTML]{2D885D}\textbf{+6.1} &
\color[HTML]{2D885D}\textbf{+9.3} &
\color[HTML]{2D885D}\textbf{+7.8} &
\color[HTML]{2D885D}\textbf{+10.3} &
\color[HTML]{2D885D}\textbf{+11.9} \\

% ================= Model: GPT-4.1 =================
\rowcolor{modelbg}
\multicolumn{17}{c}{\textbf{Model: GPT-4.1}} \\

Agentless 
& \heat{65.2} & \heat{90.8} & \heat{65.7} & \heat{63.5}
& \heat{29.3} & \heat{49.0} & \heat{32.7} & \heat{26.4}
& \heat{62.0} & \heat{85.5} & \heat{63.0} & \heat{54.5}
& \heat{35.1} & \heat{59.4} & \heat{46.0} & \heat{25.4} \\

OrcaLoca
& \heat{75.2} & \heat{80.0} & \heat{76.5} & \heat{71.3}
& \heatul{55.2} & \heat{66.7} & \heatul{59.0} & \heatul{50.1}
& \heat{56.2} & \heat{59.6} & \heat{57.1} & \heat{44.2}
& \heat{42.0} & \heat{50.5} & \heat{46.2} & \heat{29.1} \\

LocAgent
& \heatul{79.5} & \heatul{90.9} & \heatul{80.8} & \heatul{77.2}
& \heat{32.3} & \heat{65.6} & \heat{36.7} & \heat{31.2}
& \heatul{74.7} & \heatul{87.9} & \heatul{76.8} & \heatul{66.1}
& \heat{43.4} & \heat{68.7} & \heat{52.5} & \heat{38.7} \\

CoSIL
& \heat{69.8} & \heat{90.6} & \heat{70.7} & \heat{67.6}
& \heat{51.8} & \heatul{74.5} & \heat{55.3} & \heat{47.0}
& \heat{62.3} & \heat{84.7} & \heat{67.3} & \heat{55.6}
& \heatul{48.8} & \heatul{72.2} & \heatul{58.3} & \heatul{41.2} \\

\textbf{Repo-Enc}
& \heatbf{82.6} & \heatbf{93.2} & \heatbf{83.6} & \heatbf{79.3}
& \heatbf{68.7} & \heatbf{83.4} & \heatbf{71.0} & \heatbf{62.4}
& \heatbf{78.0} & \heatbf{90.5} & \heatbf{81.4} & \heatbf{69.0}
& \heatbf{64.7} & \heatbf{81.9} & \heatbf{72.1} & \heatbf{52.6} \\

\hline
\textbf{$\Delta_{\text{best}}$}
& \color[HTML]{2D885D}\textbf{+3.1}
& \color[HTML]{2D885D}\textbf{+2.3}
& \color[HTML]{2D885D}\textbf{+2.8}
& \color[HTML]{2D885D}\textbf{+2.1}
& \color[HTML]{2D885D}\textbf{+13.5}
& \color[HTML]{2D885D}\textbf{+8.9}
& \color[HTML]{2D885D}\textbf{+12.0}
& \color[HTML]{2D885D}\textbf{+12.3}
& \color[HTML]{2D885D}\textbf{+3.3}
& \color[HTML]{2D885D}\textbf{+2.6}
& \color[HTML]{2D885D}\textbf{+4.6}
& \color[HTML]{2D885D}\textbf{+2.9}
& \color[HTML]{2D885D}\textbf{+15.9}
& \color[HTML]{2D885D}\textbf{+9.7}
& \color[HTML]{2D885D}\textbf{+13.8}
& \color[HTML]{2D885D}\textbf{+11.4} \\

% ================= Model: GPT-5 =================
\rowcolor{modelbg}
\multicolumn{17}{c}{\textbf{Model: GPT-5}} \\

Agentless 
& \heat{78.7} & \heat{95.9} & \heat{78.3} & \heat{76.2}
& \heat{45.1} & \heat{68.1} & \heat{47.3} & \heat{41.3}
& \heat{64.5} & \heat{87.4} & \heat{65.1} & \heat{57.4}
& \heat{38.8} & \heat{64.6} & \heat{49.7} & \heat{31.6} \\

OrcaLoca
& \heatul{88.2} & \heat{93.9} & \heatul{88.6} & \heat{84.2}
& \heatul{76.1} & \heatul{86.2} & \heatul{79.1} & \heatul{68.6}
& \heat{74.4} & \heat{82.3} & \heat{77.6} & \heat{63.5}
& \heatul{59.6} & \heat{74.0} & \heatul{68.6} & \heat{46.6} \\

LocAgent
& \heat{88.2} & \heatul{96.7} & \heat{88.4} & \heatul{86.7}
& \heat{50.9} & \heat{80.3} & \heat{55.9} & \heat{49.7}
& \heatul{79.7} & \heatul{93.0} & \heatul{81.4} & \heatul{74.2}
& \heat{48.0} & \heat{68.7} & \heat{56.6} & \heat{40.5} \\

CoSIL
& \heat{82.8} & \heat{95.7} & \heat{82.3} & \heat{80.2}
& \heat{68.3} & \heat{81.8} & \heat{68.9} & \heat{62.3}
& \heat{69.8} & \heat{89.3} & \heat{72.9} & \heat{62.2}
& \heat{55.2} & \heatul{76.2} & \heat{62.3} & \heatul{46.5} \\

\textbf{Repo-Enc}
& \heatbf{91.9} & \heatbf{97.7} & \heatbf{91.1} & \heatbf{89.1}
& \heatbf{83.4} & \heatbf{93.6} & \heatbf{84.5} & \heatbf{76.9}
& \heatbf{82.1} & \heatbf{94.4} & \heatbf{85.4} & \heatbf{76.2}
& \heatbf{71.9} & \heatbf{87.8} & \heatbf{78.1} & \heatbf{61.1} \\

\hline
\textbf{$\Delta_{\text{best}}$}
& \color[HTML]{2D885D}\textbf{+3.7}
& \color[HTML]{2D885D}\textbf{+1.0}
& \color[HTML]{2D885D}\textbf{+2.5}
& \color[HTML]{2D885D}\textbf{+2.4}
& \color[HTML]{2D885D}\textbf{+7.3}
& \color[HTML]{2D885D}\textbf{+7.4}
& \color[HTML]{2D885D}\textbf{+5.4}
& \color[HTML]{2D885D}\textbf{+8.3}
& \color[HTML]{2D885D}\textbf{+2.4}
& \color[HTML]{2D885D}\textbf{+1.4}
& \color[HTML]{2D885D}\textbf{+4.0}
& \color[HTML]{2D885D}\textbf{+2.0}
& \color[HTML]{2D885D}\textbf{+12.3}
& \color[HTML]{2D885D}\textbf{+11.6}
& \color[HTML]{2D885D}\textbf{+9.5}
& \color[HTML]{2D885D}\textbf{+14.5} \\

% ================= Model: Claude-4.5 =================
\rowcolor{modelbg}
\multicolumn{17}{c}{\textbf{Model: Claude-4.5-Sonnet}} \\

Agentless & \heat{76.6} & \heatul{96.5} & \heat{76.9} & \heat{74.4} & \heat{31.7} & \heat{34.6} & \heat{32.0} & \heat{27.1} & \heat{63.8} & \heatul{89.7} & \heat{66.1} & \heat{58.0} & \heat{41.4} & \heat{72.4} & \heat{55.3} & \heat{35.9} \\

OrcaLoca & \heatul{87.2} & \heat{89.6} & \heatul{87.5} & \heatul{82.2} & \heatul{74.5} & \heatul{79.3} & \heatul{76.5} & \heatul{65.1} & \heatul{74.7} & \heat{78.3} & \heatul{76.2} & \heatul{61.5} & \heatul{65.1} & \heat{69.4} & \heatul{67.8} & \heatul{46.1} \\

LocAgent & \heat{71.4} & \heat{76.6} & \heat{72.7} & \heat{70.2} & \heat{49.3} & \heat{57.8} & \heat{51.5} & \heat{44.9} & \heat{58.7} & \heat{69.0} & \heat{61.6} & \heat{54.7} & \heat{47.3} & \heat{60.5} & \heat{52.6} & \heat{39.3} \\

CoSIL & \heat{75.5} & \heat{96.1} & \heat{75.9} & \heat{73.7} & \heat{57.5} & \heat{78.7} & \heat{60.7} & \heat{52.9} & \heat{64.5} & \heat{88.3} & \heat{69.4} & \heat{57.5} & \heat{51.1} & \heatul{74.9} & \heat{60.1} & \heat{39.6} \\

\textbf{Repo-Enc} & \heatbf{90.5} & \heatbf{97.6} & \heatbf{91.8} & \heatbf{88.6} & \heatbf{79.8} & \heatbf{93.7} & \heatbf{83.4} & \heatbf{75.8} & \heatbf{82.0} & \heatbf{93.9} & \heatbf{85.6} & \heatbf{75.8} & \heatbf{74.8} & \heatbf{90.4} & \heatbf{80.7} & \heatbf{63.3} \\

\hline
\textbf{$\Delta_{\text{best}}$} & 
\color[HTML]{2D885D}\textbf{+3.3} &
\color[HTML]{2D885D}\textbf{+1.1} &
\color[HTML]{2D885D}\textbf{+4.3} &
\color[HTML]{2D885D}\textbf{+6.4} &
\color[HTML]{2D885D}\textbf{+5.3} &
\color[HTML]{2D885D}\textbf{+14.4} &
\color[HTML]{2D885D}\textbf{+6.9} &
\color[HTML]{2D885D}\textbf{+10.7} &
\color[HTML]{2D885D}\textbf{+7.3} &
\color[HTML]{2D885D}\textbf{+4.2} &
\color[HTML]{2D885D}\textbf{+9.4} &
\color[HTML]{2D885D}\textbf{+14.3} &
\color[HTML]{2D885D}\textbf{+9.7} &
\color[HTML]{2D885D}\textbf{+15.5} &
\color[HTML]{2D885D}\textbf{+12.9} &
\color[HTML]{2D885D}\textbf{+17.2} \\

\bottomrule
\end{tabular}%
} % end resizebox
\end{table*}

%% file: tables/reconstruction.tex
\begin{table*}[h]
\centering
\caption{Main results on repository reconstruction tasks in RepoCraft. Gold Projects represent statistics of the original human-written repositories. More results are provided in Appendix~\ref{app:results_recon}.}
\label{tab:reconstruction_results}
\small
\setlength{\tabcolsep}{4pt}       
\resizebox{\textwidth}{!}{%
\begin{tabular}{llccccc}
\toprule
Framework & Backbone & Coverage (\%) $\uparrow$ & Accuracy (Pass / Vote) (\%) $\uparrow$ & \#Files $\uparrow$ & nLOC $\uparrow$ & Code Tokens $\uparrow$ \\
\midrule
\rowcolor{modelbg}
\multirow{1}{*}{Gold Projects (Reference)} 
  & Human Developers 
  & 100.0 & 94.8 / 98.8 
  & 345 & 97,725 & 718,946 \\
\midrule
\multirow{2}{*}{ZeroRepo-Doc (Baseline)}
  & GPT-4.1    & 64.6 & 50.0 / 63.4 & 209 & 6,079 & 158,948 \\
  & GPT-5-mini & 74.2 & 52.6 / 71.4 & 143 & 13,414 & 125,625 \\
\midrule
\multirow{2}{*}{ZeroRepo-RPG (Ours)}
  & GPT-4.1    & \textbf{93.5} & \textbf{85.8 / 93.4} & \textbf{206} & \textbf{35,190} & \textbf{346,865} \\
  & GPT-5-mini & \textbf{98.5} & \textbf{86.0 / 97.7} & \textbf{226} & \textbf{60,871} & \textbf{550,432} \\
\bottomrule
\end{tabular}%
}
\end{table*}

%% file: sections/4_ablation.tex
\section{Ablation Study}
\label{sec:abl}

\paragraph{Experimental Setup.} To isolate semantic and topological contributions, we run two ablations (detailed in Appendix~\ref{app:ablations}). For Reconstruction (RepoCraft), we progressively strip node metadata bottom-up while retaining semantic features to evaluate representational fidelity. For Understanding (SWE-bench Live), we remove structural metadata $\mathbf{m}$ to assess navigational efficacy under \ours{}.

\begin{wraptable}{r}{0.52\columnwidth}
\centering
\caption{Ablation study of \ours{} on SWE-bench Live. Best results are highlighted in \textbf{bold}.}
\label{tab:ablation_live_gpt4o_gpt41}
\renewcommand{\arraystretch}{1.15}
\setlength{\tabcolsep}{2.1pt}

\resizebox{\linewidth}{!}{%
\begin{tabular}{@{}l l cccc cccc@{}}
\toprule
\multirow{2}{*}{Backbone} & \multirow{2}{*}{Method} &
\multicolumn{4}{c}{File-level} &
\multicolumn{4}{c}{Function-level} \\
\cmidrule(lr){3-6}\cmidrule(lr){7-10}
& & Acc@1 & Acc@5 & Pre & Rec & Acc@1 & Acc@5 & Pre & Rec \\
\midrule
\multirow{3}{*}{GPT-4o}
& \ours{} & \textbf{69.2} & \textbf{83.5} & \textbf{73.2} & \textbf{60.3} & \textbf{50.5} & \textbf{69.4} & \textbf{59.4} & \textbf{41.8} \\
& w/o Dependency & 58.4 & 77.4 & 63.0 & 53.3 & 44.8 & 66.3 & 53.4 & 36.4 \\
& w/o Feature    & 60.9 & 76.3 & 64.6 & 52.4 & 43.1 & 63.4 & 52.3 & 35.5 \\
\midrule
\multirow{3}{*}{GPT-4.1}
& \ours{} & \textbf{78.0} & \textbf{90.5} & \textbf{81.4} & \textbf{69.0} & \textbf{64.7} & \textbf{81.9} & \textbf{72.1} & \textbf{52.6} \\
& w/o Dependency & 77.4 & 89.4 & 80.6 & 68.3 & 63.7 & 80.2 & 71.1 & 51.9 \\
& w/o Feature    & 71.7 & 87.5 & 76.9 & 64.5 & 57.4 & 76.3 & 66.3 & 47.8 \\
\bottomrule
\end{tabular}%
}
\end{wraptable}

\paragraph{Semantics and Topology are Mutually Reinforcing.}
Table~\ref{tab:ablation_live_gpt4o_gpt41} delineates the distinct functional contributions of graph components. Semantic Features provide essential semantic grounding for fine-grained localization; their removal causes the sharpest decline in Function-level Acc@1 ($50.5\% \to 43.1\%$ on GPT-4o), indicating that abstract summaries are indispensable for aligning natural language intent with concrete implementations. Dependencies establish structural connectivity; severing these edges disrupts execution tracing, significantly degrading File-level retrieval. The Full RPG integrates these layers to maximize context discovery, consistently outperforming all ablated variants.

\begin{wraptable}{r}{0.52\columnwidth}
\centering
\caption{Ablation on representational fidelity on \texttt{scikit-learn} (GPT-5-mini).}
\label{tab:reconstruction_ablation_sklearn_gpt5mini}
\setlength{\tabcolsep}{4.2pt}

\resizebox{\linewidth}{!}{%
\begin{tabular}{l ccccc}
\toprule
Method & Coverage & Pass Rate & \#Files & nLOC & Tokens \\
\midrule
ZeroRepo-Docs & 72.3 & 55.6 / 66.3 & 76  & 12,007 & 101,988 \\
ZeroRepo-RPG (\ours{}) & \textbf{100.0} & \textbf{82.8 / 99.5} & \textbf{256} & \textbf{96,831} & \textbf{898,026} \\
w/o Function Metadata & 91.5 & 74.1 / 90.9 & 248 & 87,413 & 854,886 \\
w/o All Node Metadata & 87.2 & 65.3 / 84.7 & 157 & 63,489 & 687,879 \\
\bottomrule
\end{tabular}%
}
\end{wraptable}

\paragraph{Hierarchical Constraints Ensure Structural Fidelity.}
Table~\ref{tab:reconstruction_ablation_sklearn_gpt5mini} indicates that the multi-level topology of RPG is essential for preserving repository modularity. Removing file and function metadata (\textit{w/o File \& Function}) results in a notable loss of structure: the number of files decreases from 256 to 157, and code volume drops by approximately 200,000 tokens. This suggests that without explicit topological boundaries, the model tends to merge distinct modules, leading to a loss of granularity. Additionally, the removal of function metadata (\textit{w/o Function Metadata}) reduces the Pass Rate from 82.8\% to 74.1\%, showing that detailed structural signals are important for code correctness. Finally, all graph-based variants outperform the text-based ZeroRepo-Docs, confirming that structured representations provide a better basis for reconstruction than linear documentation.

%% file: sections/4_analysis.tex
% \vspace*{-5pt}
\section{Analysis}
\label{sec:analysis}
\subsection{Representational Efficiency}

\input{tables/cost}
\paragraph{RPG Facilitates Reasoning Efficiency.}
Table~\ref{tab:efficiency_analysis} evaluates the efficiency of agents guided by different substrates. Across all backbones, \ours{} achieves fewer steps and lower expenditure, yielding the highest cost-effectiveness (Acc@5/Cost). On GPT-5, \ours{} reaches an efficiency of 4.15 at a cost of \$0.22, whereas baselines such as OrcaLoca and LocAgent require higher expenditures for lower efficiency gains. This trend is consistent with GPT-4.1 results, where \ours{} attains the peak efficiency of 4.63. These results indicate that RPG-guided navigation enables precise exploration, concentrating reasoning resources on relevant code regions and reducing redundant API calls throughout the localization process.
% \vspace*{-5pt}
\subsection{Structural Evolvability}

\begin{figure}[htbp]
\vspace{0pt}
\centering

\begin{minipage}[t]{0.49\linewidth}
\vspace{0pt} 
\centering
\includegraphics[width=\linewidth]{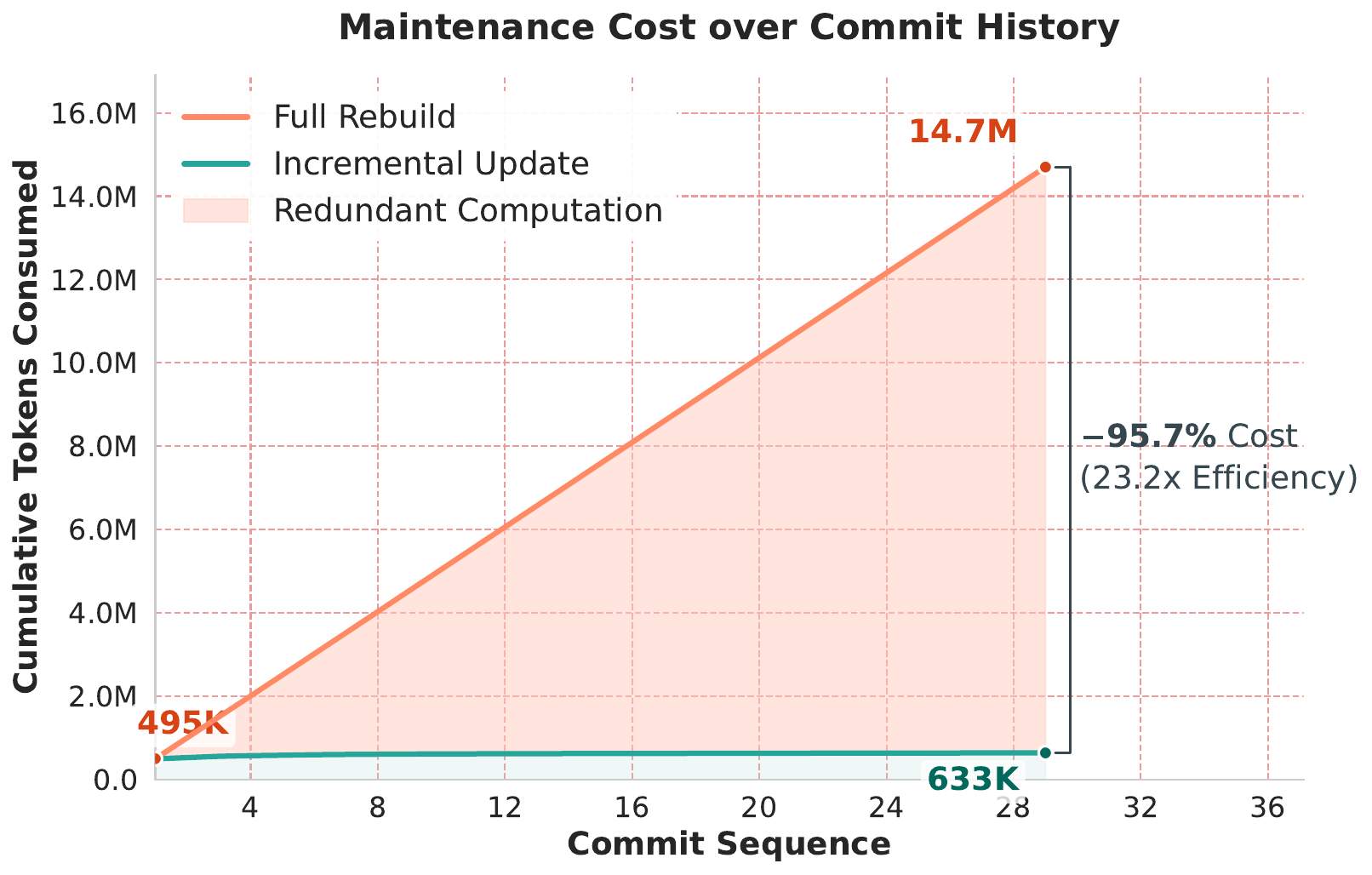}
\caption{Cost Efficiency Comparison: RPG Rebuilding versus Incremental Updates across Commit History.}
\label{fig:maintenance_cost}
\end{minipage}
\hfill
\begin{minipage}[t]{0.49\linewidth}
\vspace{15pt} 
\centering
\captionof{table}{Full vs. Incremental RPG Fidelity on SWE-bench Live. SWE-bench Live accuracy of RPGs across commits under full reconstruction (Full) and incremental maintenance (Incr.).}
\label{tab:full_incr_fidelity}
\footnotesize
\renewcommand{\arraystretch}{1.02}
\setlength{\tabcolsep}{3.0pt}

\resizebox{\linewidth}{!}{%
\begin{tabular}{@{}l l cccc cccc@{}}
\toprule
\multirow{2}{*}{Model} & \multirow{2}{*}{Strategy} &
\multicolumn{4}{c}{File-level} &
\multicolumn{4}{c}{Function-level} \\
\cmidrule(lr){3-6}\cmidrule(lr){7-10}
& & Acc@1 & Acc@5 & Pre & Rec & Acc@1 & Acc@5 & Pre & Rec \\
\midrule
\multirow{2}{*}{GPT-4o}
& Full
& \textbf{69.9} & \textbf{84.6} & \textbf{73.2} & 60.1
& \textbf{53.8} & 68.5 & \textbf{60.6} & 41.1 \\
& Incr.
& 69.2 & 83.5 & \textbf{73.2} & \textbf{60.3}
& 50.5 & \textbf{69.4} & 59.4 & \textbf{41.8} \\
\midrule
\multirow{2}{*}{GPT-4.1}
& Full
& \textbf{79.9} & 88.2 & \textbf{82.5} & \textbf{69.8}
& \textbf{67.4} & 80.3 & \textbf{73.3} & \textbf{55.4}  \\
& Incr.
& 78.0 & \textbf{90.5} & 81.4 & 69.0
& 64.7 & \textbf{81.9} & 72.1 & 52.6 \\
\bottomrule
\end{tabular}%
}
\end{minipage}

\vspace{-10pt}
\end{figure}

\paragraph{Incremental Maintenance Ensures Sustainable Scalability.}
To assess feasibility, we measure maintenance costs across a commit sequence. Figure~\ref{fig:maintenance_cost} shows that full reconstruction scales linearly and exceeds 14.7M tokens, whereas our incremental strategy uses only 633K tokens by isolating semantic deltas. This 95.7\% reduction confines heavy computation to a one-time initialization and effectively decouples ongoing maintenance costs from repository scale, enabling sustainable long-term operation.

\begin{figure*}[t]
    \centering
    \includegraphics[width=\linewidth]{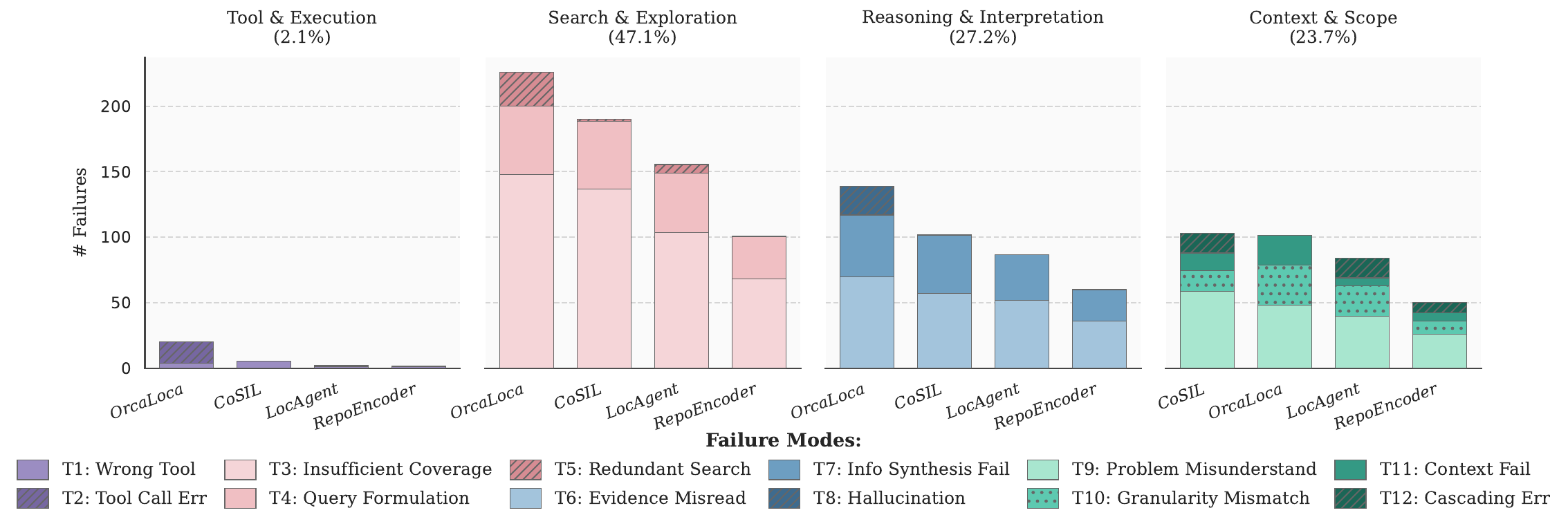}
    \caption{Distribution of Failure Modes on SWE-bench Verified. We analyze 100 failed trajectories per method with GPT-4o. Errors fall into four macro-groups: Tool \& Execution, Search \& Exploration, Reasoning \& Interpretation, and Context \& Scope, with 12 sub-types (T1–T12). See Appendix~\ref{app:err_analysis}.}
    \label{fig:error_dist}
    % \vspace*{-5pt}
\end{figure*}

\paragraph{Evolution Balance between Fidelity and Efficiency.} To validate resilience against semantic drift during updates, we assessed representational fidelity by deploying agents on SWE-bench Live using RPGs from both strategies. Table~\ref{tab:full_incr_fidelity} indicates that the "Incr." strategy maintains statistical parity with the "Full" baseline. Specifically, while "Incr." achieves slightly higher retrieval accuracy (81.9\% Acc@5 compared to 80.3\% for GPT-4.1), "Full" reconstruction retains a marginal edge, surpassing "Incr." by approximately 2\% in Precision and Recall. This balance confirms that our sustainable evolution effectively preserves the repository's semantic integrity with negligible degradation.

% \vspace*{-2pt}
\subsection{Agentic Navigability}

\begin{wrapfigure}[18]{r}{0.52\columnwidth}
    \centering
    \includegraphics[width=\linewidth]{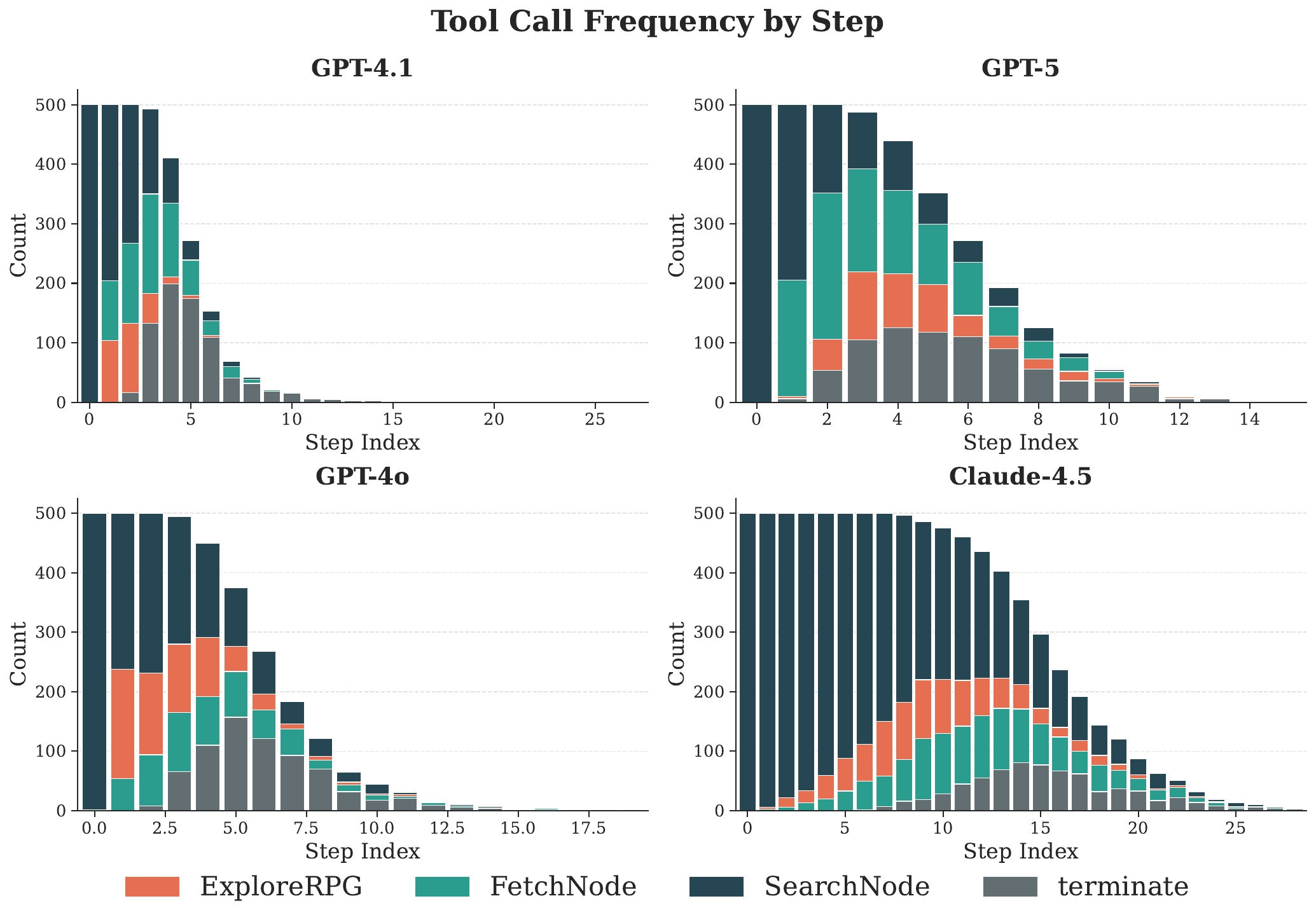}
    \caption{Impact of RPG Tooling on Agent Behavior on SWE-bench Verified. Step-wise action distributions induced by the RPG interface across LLMs.}
    \label{fig:agent_step}
    \vspace*{-10pt}
\end{wrapfigure}

\paragraph{RPG Induces Structured Exploration.}
To investigate whether RPG structures reasoning, we visualized tool usage distributions across LLMs. Figure~\ref{fig:agent_step} reveals a universal "Search-then-Zoom" pattern: agents prioritize broad topology traversal (ExploreRPG, SearchNode) to establish a global map before narrowing to fine-grained analysis (FetchNode). This trend is more pronounced in stronger reasoners (e.g., Claude-4.5), which leverage RPG's structural context to support extended interaction horizons. These results confirm that RPG effectively guides agents from global comprehension to localized implementation.

\paragraph{Dual-View Search Mitigates Navigational Failures.}
We manually analyzed 100 failed trajectories from GPT-4o to identify error patterns mitigated by the RPG structure. As shown in Figure~\ref{fig:error_dist}, RPG reduces Search \& Exploration failures compared to baselines. While systems like LocAgent and CoSIL utilize graph structures, they often suffer from Insufficient Coverage. \ours{} addresses this by providing dual-path access, where semantic features enable broad global retrieval to expand the search space, while the structured hierarchy guides the agent to reduce Redundant Search. This multi-view navigation ensures agents can accurately localize intent before traversing implementation-level dependencies. Improved localization also reduces downstream errors in Context \& Scope, keeping reasoning grounded in the correct implementation units.

%% file: tables/cost.tex
\begin{wraptable}{r}{0.48\textwidth}
\vspace{-6pt}
\centering
\caption{Efficiency for repository understanding on SWE-bench Verified. Steps and Cost are averaged over tasks. Eff. is defined as $\mathrm{Acc@5}/\mathrm{Cost}$. Additional results are provided in Appendix~\ref{app:cost_analysis}.}
\label{tab:efficiency_analysis}
\footnotesize
\setlength{\tabcolsep}{4pt}

\resizebox{\linewidth}{!}{%
\begin{tabular}{l c c c c c c}
\toprule
\multirow{2}{*}{Method}
& \multicolumn{3}{c}{GPT-4.1}
& \multicolumn{3}{c}{GPT-5} \\
\cmidrule(lr){2-4}\cmidrule(lr){5-7}
& Steps & Cost (\$) & Eff.
& Steps & Cost (\$) & Eff. \\
\midrule
OrcaLoca
& 20.22 & 0.46 & 1.48
& 36.93 & 0.75 & 1.16 \\
CoSIL
& 19.77 & 0.24 & 3.10 
& 19.52 & 0.31 & 2.64 \\
LocAgent
& 11.94 & 0.86 & 0.76
&  6.48 & 0.49 & 1.64 \\
\ours{}
& \textbf{6.75} & \textbf{0.18} & \textbf{4.63} 
& \textbf{6.34} & \textbf{0.22} & \textbf{4.15} \\
\bottomrule
\end{tabular}%
}
\vspace{-10pt}
\end{wraptable}

% % ================= Panel B =================
% \begin{table}[htbp]
% \centering
% \caption{Overall efficiency analysis (\textbf{Panel B}): Repository Reconstruction (Representational Efficiency).}
% \label{tab:efficiency_panel_b}
% \footnotesize
% \setlength{\tabcolsep}{4pt}

% \resizebox{\columnwidth}{!}{%
% \begin{tabular}{l c}
% \toprule
% \textbf{Representation} & \textbf{Input Tokens / Reduction} \\
% \midrule
% API Doc & 482,306 \quad / \quad -- \\
% \textbf{RPG (Ours)} & \textbf{55,060} \quad / \quad \textbf{-88.6\%} \\
% \bottomrule
% \end{tabular}%
% }
% \end{table}

%% file: sections/5_conclusion.tex
\section{Conclusion}
In this work, we introduce \ours{}, transforming the Repository Planning Graph (RPG) into a unified representation for repository reasoning. By coupling dense semantics with topological constraints, \ours{} bridges architectural intent and implementation. Our evaluations show RPG is a superior navigational map for localization and a blueprint for reconstruction, achieving significantly higher fidelity than documentation. Furthermore, our incremental mechanism ensures consistency with lower overhead. Ultimately, \ours{} establishes a robust foundation for closed-loop software engineering by bidirectionally linking architectural intent with structural implementation.

%% file: sections/appendix.tex
\appendix

\addtocontents{toc}{\protect\setcounter{tocdepth}{3}}
\renewcommand{\contentsname}{Appendix Contents for \ours{}}
\hypersetup{linkcolor=black}
\tableofcontents 
\hypersetup{linkcolor=red}
\clearpage

\lstset{
    basicstyle=\footnotesize\ttfamily,
    breaklines=true,
    frame=lines,  
    breakindent=0pt,
    extendedchars=true,
    belowcaptionskip=0.5em,
    escapechar=@,
    literate={á}{{\'a}}1 {ã}{{\~a}}1 {é}{{\'e}}1 {£}{{\pounds}}1 {–}{{-}}1 {’}{{'}}1,
}

\input{appendix/method}

\input{appendix/exp_setup}

\input{appendix/results}
\input{appendix/ablations}

%% file: appendix/method.tex
\section{Detailed Methodology of \ours{}}
\label{app:rpg_encoder}

This section provides a deep dive into the implementation details and algorithmic foundations of \ours{}, expanding upon the three core phases—Construction, Evolution, and Operation—introduced in Section 3 of the main text.

\input{appendix/methods/extraction}

\input{appendix/methods/evolution}

\input{appendix/methods/operation}

%% file: appendix/methods/extraction.tex
% =========================
% A.1 RPG Extraction
% =========================
\subsection{RPG Extraction: Semantic Lifting and Hierarchical Encoding}
\label{app:extraction}

This subsection details the construction stage of RPG-Encoder, which transforms a raw repository into a
hierarchically organized \emph{feature space} (the semantic backbone of RPG) together with a \emph{grounded}
mapping that links abstract functional nodes to concrete directory scopes.
Concretely, the extraction stage proceeds in three steps:
(1) \textbf{Semantic lifting} that converts low-level code entities into atomic functional features;
(2) \textbf{Latent architecture recovery} that reorganizes these features into a consistent three-level hierarchy;
and (3) \textbf{Artifact grounding} that anchors each abstract subtree to a compact set of physical directory paths.
The resulting hierarchy serves as the \emph{Functionality SubGraph} used by downstream agentic tools (Appendix~\ref{app:operation}).

% -------------------------
% A.1.1 Semantic Lifting
% -------------------------
\subsubsection{Semantic Lifting via Prompted Semantic Parsing}
\label{app:extraction_semantic}

\paragraph{Global parsing strategy.}
Given a repository $\mathcal{R}$, semantic lifting is performed from a \emph{global perspective} rather than on
individual files in isolation. We first identify all code entities of interest, including classes, methods and functions, and treat them as the fundamental semantic units to be analyzed.
This global view allows the model to maintain consistent semantic granularity across the repository and reduces
local biases introduced by file boundaries.

\paragraph{Semantic units and batching.}
To accommodate repositories of varying scales while respecting model context limits, code entities are abstracted into \emph{semantic units} and analyzed in batches under a controlled token budget. Each semantic unit represents a coherent functional entity,
ensuring that semantically coupled components are interpreted in context.
Batches are constructed to balance completeness and efficiency, such that every semantic unit is analyzed exactly once while enabling scalable processing of large repositories.

\paragraph{Semantic feature representation.}
For each code entity $u$, the parser produces a set of \emph{atomic semantic features}
$f(u)=\{a_1, a_2, \dots\}$, where each $a_i$ is a short verb--object phrase describing \emph{what} the entity does rather than \emph{how} it is implemented. These atomic features are intentionally constrained to be:
(i) \textbf{single-responsibility}, (ii) \textbf{implementation-agnostic}, and (iii) \textbf{lexically normalized}
(lowercase English, concise phrasing). This normalization is critical for subsequent routing and hierarchical encoding,
since it provides stable semantic anchors for grouping and comparison across the repository.

\paragraph{Prompt template (semantic parsing).}
We implement semantic lifting using the following prompt template, which enforces: (1) complete coverage of all
functions in the chunk, (2) strict output schema, and (3) feature naming rules that avoid vague verbs and
implementation details. The prompt returns a JSON object mapping each function name to a list of semantic features.

\input{prompts/semantic_parsing}

\paragraph{Post-processing and validation.}
We apply lightweight validation to guarantee the output is machine-consumable:
(i) JSON parsing and schema checking (every function in the input must appear as a key);
(ii) feature list normalization (whitespace, casing, deduplication);
and (iii) optional merging for decorator variants (e.g., property getter/setter) when they share the same method name,
as allowed by the prompt. If the model returns malformed output, we retry with a minimal format correction instruction
without changing semantic constraints.

\begin{figure}[htbp]
    \centering
    \includegraphics[width=\linewidth]{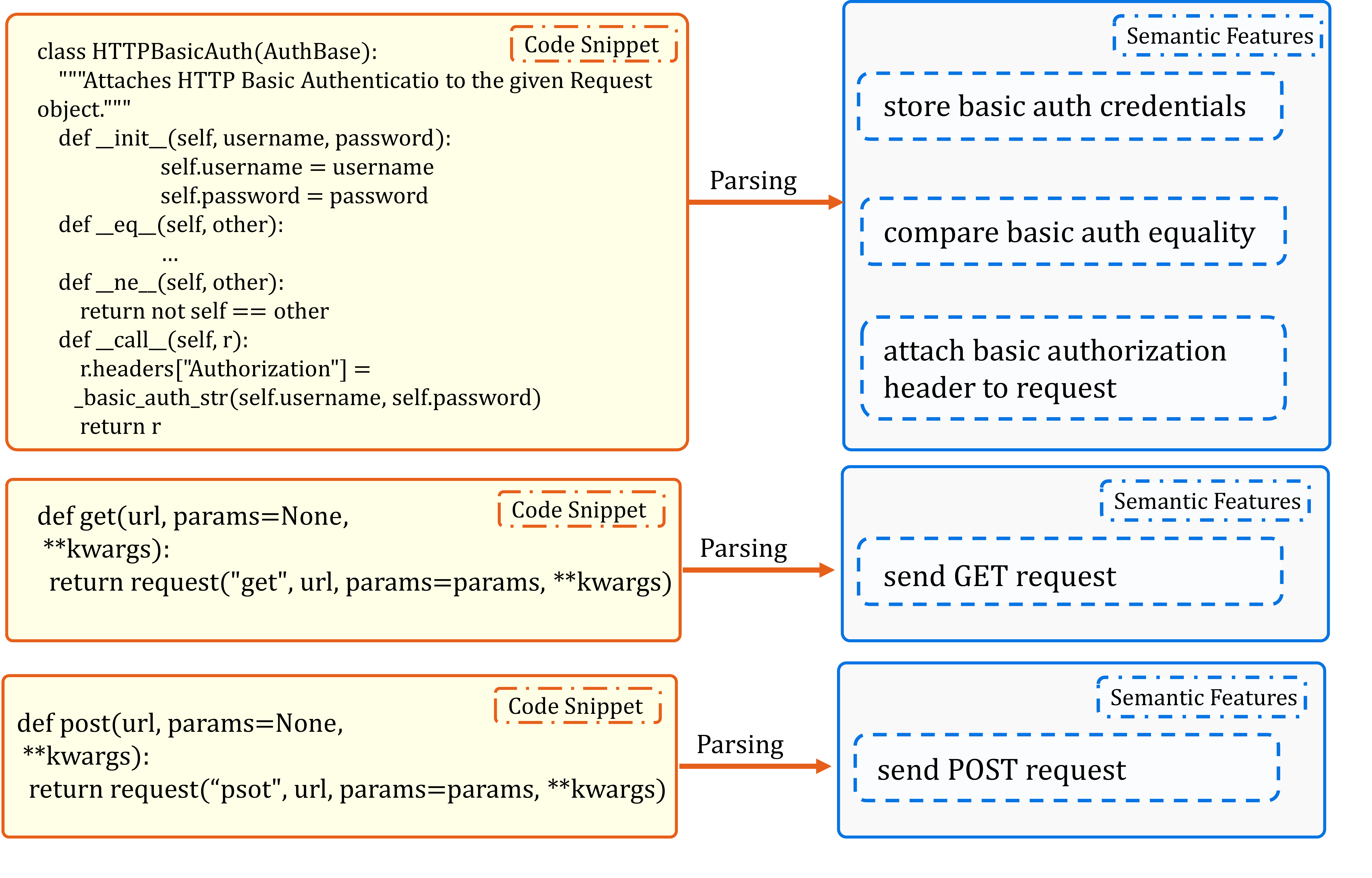}
    \caption{Illustration of raw code snippets and their corresponding semantic features extracted via semantic parsing.}
    \label{fig:parsing_process}
\end{figure}

\paragraph{Illustrative example.}
Figure~\ref{fig:parsing_process} shows an end-to-end example of semantic lifting, where raw code snippets are mapped
to their corresponding atomic semantic features.

% -------------------------
% A.1.2 Latent Architecture Recovery
% -------------------------
\subsubsection{Latent Architecture Recovery for Hierarchical Encoding}
\label{app:extraction_latent}

\paragraph{Motivation.}
Semantic lifting yields a set of fine-grained features distributed across many files, which is insufficient as a
planning substrate: flat features are hard to navigate, while directory-only grouping often overlooks logical roles.
We therefore recover a \emph{latent functional architecture} that reorganizes the repository into a consistent,
interpretable, and searchable hierarchy. We enforce a strict \textbf{three-level} feature path format:
\[
\texttt{<functional area>/<category>/<subcategory>},
\]
which balances abstraction (top-level intent) and specificity (fine-grained specialization), while keeping routing and
tool-based navigation tractable.

\paragraph{Step 1: Domain discovery (functional areas).}
We first discover a small set of high-level functional areas that act as architectural centroids. The model is guided
to propose meaningful areas (e.g., \texttt{DataProcessing}, \texttt{ModelTraining}, \texttt{EvaluationMetrics}) while
avoiding low-signal directories such as vendor code, tests, or documentation.

\input{prompts/domain_discovery}

\paragraph{Step 2: Hierarchical construction (three-level paths).}
Given the discovered functional areas and the parsed feature groups, we perform hierarchical construction by assigning
each top-level feature group to a unique three-level target path. This step is formulated as a constrained semantic
assignment problem: the model must use only the provided functional areas for the first level, and it must generate
intent-focused category/subcategory labels following the same semantic naming rules used in semantic lifting.

\input{prompts/hierarchical_construction}

\paragraph{Outputs and usage.}
The output of hierarchical construction is a mapping from feature paths to sets of feature groups, which induces a
topological feature tree $\mathcal{T}_{\text{feature}}$. This tree serves two purposes:
(i) it provides high-signal \emph{search scopes} for intent-to-code mapping, and
(ii) it supports routing and traversal by ensuring semantically coherent boundaries at each level.

\begin{figure}[htbp]
    \centering
    \includegraphics[width=\linewidth]{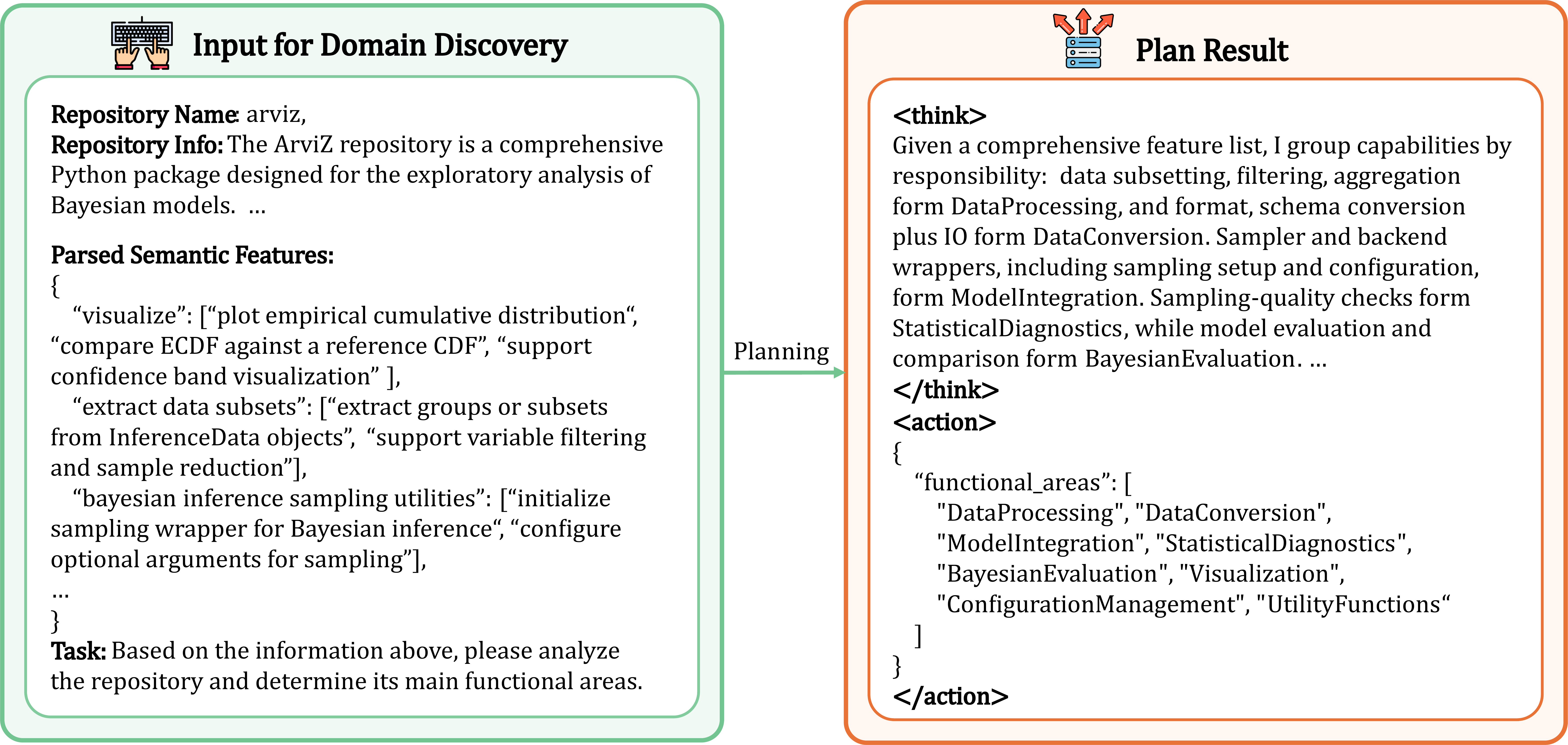}
    \caption{Illustrative example of the Domain Discovery phase.}
    \label{fig:domain_discovery_example}
\end{figure}

\begin{figure}[htbp]
    \centering
    \includegraphics[width=\linewidth]{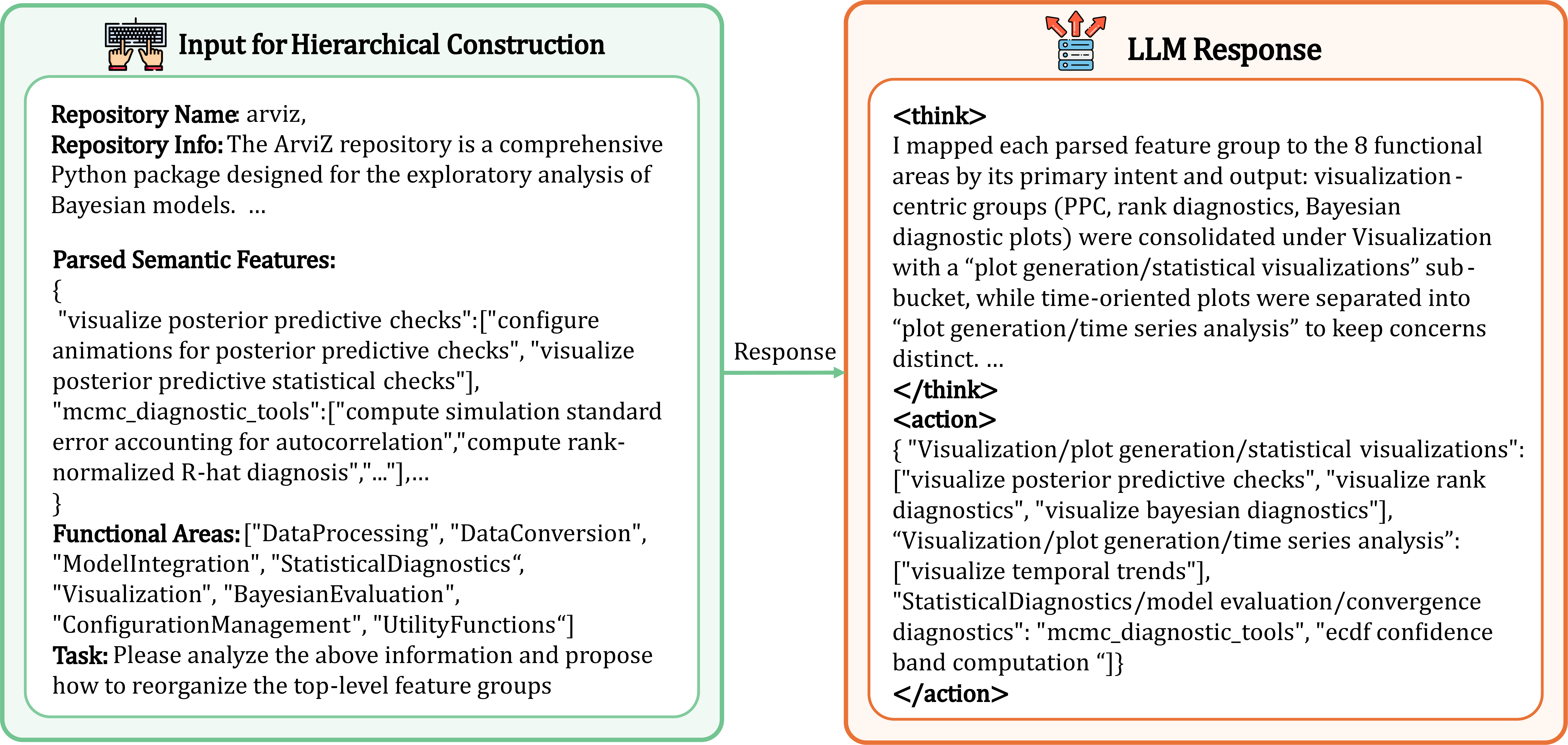}
    \caption{Example of the Hierarchical Construction phase.}
    \label{fig:hierarchical_construction}
\end{figure}

\paragraph{Illustrative examples.}
Figures~\ref{fig:domain_discovery_example} and~\ref{fig:hierarchical_construction} provide examples of domain discovery
and hierarchical construction, respectively.

% -------------------------
% A.1.3 Artifact Grounding
% -------------------------
\subsubsection{Artifact Grounding: Anchoring Abstract Subtrees to Directory Scopes}
\label{app:extraction_grounding}

\paragraph{Problem formulation.}
To bridge the semantic hierarchy $\mathcal{T}_{\text{feature}}$ with physical repository artifacts, we ground each
abstract node $v$ to a compact set of directory scopes.
Let $\mathcal{L}(v)$ denote the set of leaf nodes in the subtree rooted at $v$.
For each leaf node $l \in \mathcal{L}(v)$, let $\text{path}(l)$ be its physical file path.
We define the \textbf{File Coverage} $\mathcal{C}(v)$ as the collection of parent directories for all leaves under $v$:
\begin{equation}
\mathcal{C}(v) = \{ \text{dir}(\text{path}(l)) \mid l \in \mathcal{L}(v) \},
\end{equation}
where $\text{dir}(\cdot)$ extracts the directory component of a file path.
We seek a compact representation $\hat{\pi}(v)$ that succinctly covers $\mathcal{C}(v)$ while preserving functional
boundaries across distinct modules.

\paragraph{Bottom-up propagation with Trie-based branching analysis.}
A naive common-prefix (LCA) computation may over-collapse unrelated modules into overly general roots (e.g., \texttt{/}).
To avoid this, we compute $\hat{\pi}(v)$ via a bottom-up propagation strategy that aggregates coverage and then
simplifies it using a Trie-based branching analysis: all paths in $\mathcal{C}(v)$ are inserted into a Prefix Tree,
and only \textbf{branching nodes} (multiple children or path termination) are retained as grounded scopes.
This yields a minimal set of directory LCAs that dominate disjoint coverage regions while respecting module boundaries.
Algorithm~\ref{alg:metadata-propagation} provides the full procedure.

\input{algos/meta_prog}

\paragraph{Complexity analysis.}
Let $N$ be the number of paths and $L$ the maximum directory depth.
Trie construction and branching-node extraction take $O(N\cdot L)$ time, bounded by total path characters.
Since propagation visits each feature node once in a bottom-up pass, the total grounding overhead scales linearly with
repository size and is negligible compared with LLM inference.

%% file: prompts/semantic_parsing.tex
\begin{tcblisting}{
  title={Semantic Parsing Prompt},
  colback=lightgray,
  colframe=black,
  arc=1mm,
  boxrule=1pt,
  left=1mm,right=1mm,top=1mm,bottom=1mm,
  breakable,
  fontupper=\scriptsize\ttfamily,
  listing only,
  listing engine=listings,
  listing options={
    breaklines=true,
    breakatwhitespace=false,
    breakindent=0pt,
    prebreak=\mbox{},
    postbreak=\mbox{},
    keepspaces=true,
    columns=fullflexible,
    tabsize=4
  }
}
## Instruction
You are a senior software analyst.
Your goal is to analyze all functions in the current input and return their key semantic features -- what each function does, not how it's implemented.

### Key Goals
- Complete analysis: Provide semantic feature extraction for every function in the given input. Do not skip any function.
- Batch perspective: Analyze all functions in the chunk together, considering their roles within the overall system.
- High-level behavior: Focus on the purpose and role of each function, not on low-level implementation details.
- If multiple definitions share the same method name (e.g., property getter and setter for the same attribute), you may output that method name only once and merge their semantic features; you do not need to distinguish decorator variants.

## Feature Extraction Principles
Follow these principles when analyzing functions:
1. Focus on the purpose and behavior of the function -- what role it serves in the system.
2. Do NOT describe implementation details, variable names, or internal logic such as loops, conditionals, or data structures.
3. If a function performs multiple responsibilities, break them down into separate features.
4. Use your understanding of each function's name, signature, and code to infer its intent.
5. Only analyze functions included in the current input -- do not guess or invent other functions.
6. Do not omit any function, including utility or helper functions.

### Feature Naming Rules:
1. Use verb + object format (e.g., `load config`, `validate token`).
2. Use lowercase English only.
3. Describe purpose not implementation (focus on what, not how).
4. Each feature must express one single responsibility.
5. If a method has multiple responsibilities, split into multiple atomic features.
6. Keep features short and atomic (prefer 3--8 words; no full sentences; no punctuation).
7. Avoid vague verbs (`handle`, `process`, `deal with`); prefer precise verbs (`load`, `validate`, `convert`, `update`, `serialize`, `compute`, `check`, `transform`).
8. Avoid implementation details (no loops, conditionals, data structures, control flow).
9. Avoid libraries/frameworks/formats (say `serialize data`, not `pickle object` / `save to json`).
10. Prefer domain/system semantics over low-level actions (`manage session` > `update dict`).
11. Avoid chaining actions (don't write `initialize config and register globally`; split into separate features).

## Output Format
You must respond with the following structure:
A `<solution>` block -- a JSON object mapping each function name to a list of its semantic features.
If a function does not implement any meaningful features (e.g., it's a stub), still include it with an empty list.

### Output Template:
<solution>
{{
  "func_name_1": ["feature one", "feature two"],
  "func_name_2": [],
  ...
}}
</solution>

## Input Context
### Repository Name
<repo_name>
{repo_name}
</repo_name>

### Repository Overview
<repo_info>
{repo_info}
</repo_info>
\end{tcblisting}

%% file: prompts/domain_discovery.tex
\begin{tcblisting}{
  title={Domain Discovery Prompt},
  colback=lightgray,
  colframe=black,
  arc=1mm,
  boxrule=1pt,
  left=1mm,right=1mm,top=1mm,bottom=1mm,
  breakable,
  fontupper=\scriptsize\ttfamily,
  listing only,
  listing engine=listings,
  listing options={
    breaklines=true,
    breakatwhitespace=false,
    breakindent=0pt,
    prebreak=\mbox{},
    postbreak=\mbox{},
    keepspaces=true,
    columns=fullflexible,
    tabsize=4
  }
}
## Instructions
You are an expert software architect and repository analyst.
Your goal is to analyze the repository holistically and identify its main functional areas -- coherent, high-level modules or subsystems that reflect the repository's architecture and purpose.

### Guidelines
- Think from a software architecture perspective; group code into major, distinct responsibilities (e.g., data loading/processing, training/inference, evaluation/metrics, visualization/reporting, APIs/interfaces, configuration/utilities/infrastructure).
- Avoid listing individual files or small helpers, third-party/vendor code, and build/test/docs directories.
- Ensure each area is meaningful and represents a clear responsibility in the codebase.

### Naming Principles
- Single Responsibility: Each area should cover one logical concern (e.g., "DataProcessing", "ModelTraining").
- High-Level Abstraction: Group related submodules; separate distinct layers.
- Consistency: Use PascalCase for names (e.g., "FeatureExtraction", "EvaluationMetrics").
- Meaningful Scope:
  - Merge closely related components (e.g., "data_loader", "dataset" -> "DataProcessing")
  - Avoid vague terms like "core", "misc", "other"
  - Use domain-specific names when appropriate (e.g., "TextPreprocessing", "ImageSegmentation")

### Output Format
Return only the result in this exact format:
<solution>
[
"functional_area1", "functional_area2", "functional_area3", ...
]
</solution>
\end{tcblisting}

%% file: prompts/hierarchical_construction.tex
\begin{tcblisting}{
  title={Hierarchical Construction Prompt},
  colback=lightgray,
  colframe=black,
  arc=1mm,
  boxrule=1pt,
  left=1mm,right=1mm,top=1mm,bottom=1mm,
  breakable,
  fontupper=\scriptsize\ttfamily,
  listing only,
  listing engine=listings,
  listing options={
    breaklines=true,
    breakatwhitespace=false,
    breakindent=0pt,
    prebreak=\mbox{},
    postbreak=\mbox{},
    keepspaces=true,
    columns=fullflexible,
    tabsize=4
  }
}
## Instruction
You are an expert software architect and large-scale repository refactoring specialist.

## Goal
Reorganize and enrich the repository's parsed feature tree by assigning each top-level feature group
(e.g., "data_loader", "model_trainer", "metrics") to the most semantically appropriate location
within the target architecture.

## Target Path Format (STRICT)
Each target path must have exactly three levels:
`<functional_area>/<category_level_1>/<subcategory_level_2>`
- `functional_area` must be one of the provided <functional_areas>.
- `category_level_1` expresses broader purpose or lifecycle role.
- `subcategory_level_2` adds precise specialization or context.
- Each segment: concise (2--5 words), semantically meaningful, intent-focused.
Examples:
- "data ingestion/pipeline orchestration/task scheduling"
- "model training/optimization strategy/hyperparameter tuning"
Avoid filler labels (e.g., "misc", "others", "core", "general").

## Semantic Naming Rules
When creating or adjusting semantic labels (categories/subcategories), follow:
1. Use "verb + object" phrasing; e.g., `load config`, `validate token`.
2. Use lowercase English only.
3. Describe purpose, not implementation.
4. Ensure each label expresses a single responsibility.
5. When multiple distinct roles exist, use multiple precise labels rather than one overloaded label.
6. Avoid vague verbs such as `handle`, `process`, and `deal with`.
7. Avoid implementation details, including control-flow or data-structure references.
8. Avoid mentioning specific libraries, frameworks, or formats; prefer `serialize data` over `pickle object` or `save to json`.
9. Prefer domain or system semantics over low-level actions; use `manage session` rather than `update dict`.

## Scope Constraints
- Only assign top-level groups (keys of <parsed_folder_tree>).
- Exclude docs/examples/tests/vendor code unless essential to core functionality.
- Do not invent new functional areas; use only those in <functional_areas>.
- You may define new categories/subcategories as needed, but they must remain meaningful and consistent.

## Output Format (STRICT)
Return only the JSON object wrapped exactly as:
<solution>
{
  "<functional_area>/<category>/<subcategory>": ["top_level_group_1", "top_level_group_2", ...],
  "<functional_area>/<category>/<subcategory>": ["top_level_group_3", ...]
}
</solution>
\end{tcblisting}

%% file: algos/meta_prog.tex
\begin{algorithm}[h]
\caption{Bottom-Up Path Metadata Propagation}
\label{alg:metadata-propagation}
\begin{algorithmic}[1]
\Require Feature Tree $T = (V, E)$, Leaf paths $\text{path}(\cdot)$
\Ensure Grounded path assignments $\hat{\pi}(v)$ for all $v \in V$

\State \textbf{Function} \textsc{Propagate}$(v)$:
    \If{$v$ is Leaf}
        \State \Return $\{ \text{dir}(\text{path}(v)) \}$ \Comment{Base case: Return physical directory}
    \EndIf
    
    \State $\mathcal{S} \leftarrow \emptyset$
    \For{$child \in \text{Children}(v)$}
        \State $\mathcal{S} \leftarrow \mathcal{S} \cup \textsc{Propagate}(child)$ \Comment{Recursively aggregate child coverage}
    \EndFor
    
    \State $\hat{\pi}(v) \leftarrow \textsc{ComputeLCA}(\mathcal{S})$ \Comment{Abstract concrete paths into logical scopes}
    \State \Return $\mathcal{S}$ \Comment{Propagate full coverage to upper layers}
\State \textbf{End Function}

\Statex
\State \textbf{Function} \textsc{ComputeLCA}$(\mathcal{S})$:
    \State $Trie \leftarrow \textsc{BuildTrie}(\mathcal{S})$ \Comment{Construct Prefix Tree from path set}
    \State $L \leftarrow \emptyset$
    \For{$node \in \textsc{PostOrder}(Trie)$} \Comment{Bottom-up traversal for optimal pruning}
        \If{$node.\text{is\_branching}()$ \textbf{or} $node.\text{is\_terminal}()$}
             \State $L.\text{add}(node.\text{path})$ \Comment{Identify meaningful functional boundary}
             \State \textsc{PruneSubtree}$(node)$ \Comment{Consolidate redundant sub-paths}
        \EndIf
    \EndFor
    \State \Return $L$
\State \textbf{End Function}
\end{algorithmic}
\end{algorithm}

%% file: appendix/methods/evolution.tex
% =========================
% A.2 Incremental Evolution
% =========================
\subsection{Incremental Evolution: Differential Update and Maintenance}
\label{app:evolution}

This subsection details how RPG-Encoder maintains the Repository Planning Graph (RPG) under continuous codebase
evolution. Given a repository update (e.g., a commit), our goal is to \emph{incrementally} update the semantic
hierarchy and its grounded mapping, ensuring the RPG remains a faithful semantic reflection of the codebase while
avoiding expensive full reconstruction. We formulate repository evolution as a stream of \textbf{atomic operations}:
\textsc{Delete}, \textsc{Modify}, and \textsc{Insert}. Each operation updates both (i) the local semantic
representation of affected entities and (ii) their placement within the feature hierarchy.

% -------------------------
% A.2.1 Differential Event Detection
% -------------------------
\subsubsection{Differential Event Detection and Operation Scheduling}
\label{app:evolution_diff}

\paragraph{From code diffs to semantic events.}
Given a code change $\Delta$ (e.g., a git diff between two revisions), we extract changed code entities at the
function/method granularity whenever possible. Each affected entity $u$ is categorized into one of three evolution events:
\begin{itemize}
    \item \textbf{Deletion:} $u$ is removed from the repository.
    \item \textbf{Modification:} $u$ exists in both revisions but its implementation changes.
    \item \textbf{Insertion:} $u$ is newly introduced in the new revision.
\end{itemize}
For \textsc{Modification}, we further distinguish between semantically stable edits and substantial semantic drift (Section~\ref{app:evolution_modify}), which determines whether the update can be handled locally or requires
structural relocation.

\paragraph{Scheduling principle.}
We schedule evolution operations under constraints that preserve structural consistency of the hierarchy,
prevent intermediate abstract nodes from accumulating dead branches,
and ensure that newly introduced entities do not disrupt the existing topological organization.

% -------------------------
% A.2.2 Deletion
% -------------------------
\subsubsection{Node Deletion with Structural Hygiene}
\label{app:evolution_delete}

\input{algos/deletion}

\paragraph{Motivation.}
Deletion must maintain structural integrity of the hierarchy. Removing a leaf entity may render its ancestor abstract
nodes semantically vacuous (i.e., nodes that no longer cover any concrete code entities). Without cleanup, these dead
branches accumulate and reduce the signal-to-noise ratio for search and routing.

\paragraph{Recursive pruning.}
We enforce \textbf{structural hygiene} via bottom-up pruning: after removing a leaf node, we recursively delete any
ancestor abstract node whose subtree becomes empty. Pruning terminates once an ancestor still has remaining children
or once the root is reached.
This mechanism prevents stale semantic categories from persisting after refactors and ensures that the hierarchy remains compact and representative of the current repository state.

\paragraph{Algorithm.}
Algorithm~\ref{alg:deletion} specifies the deletion procedure and the recursive pruning logic.

% -------------------------
% A.2.3 Modification
% -------------------------
\subsubsection{Differential Modification Processing}
\label{app:evolution_modify}

\input{algos/modified}

\paragraph{Motivation.}
A code edit may either preserve the original intent (e.g., bug fixes, refactoring, parameter tuning) or substantially
change functionality (semantic drift). Treating both cases identically is suboptimal: in-place updates are sufficient
for minor edits, while major drift requires relocating the entity to a semantically congruent domain.

\paragraph{Minor update vs.\ semantic drift.}
Given a modified entity $u$ with old/new versions $(u^{old}, u^{new})$, we compute semantic features
$f(u^{old})$ and $f(u^{new})$ using the same parsing constraints as in extraction. We then assess drift based on:
(i) feature overlap/consistency, and (ii) an LLM judgement constrained by explicit criteria. If drift is minor, we perform an in-place update
of the node’s semantic summary; otherwise, we trigger re-routing.

\paragraph{Re-routing as composition.}
When semantic drift is significant, we treat modification as a composition of atomic operations:
\[
\textsc{Modify}(u) \;\Rightarrow\; \textsc{Delete}(u^{old}) + \textsc{Insert}(u^{new}),
\]
which relocates the entity to a new functional domain via the same semantic routing procedure used for insertion.
This guarantees that the hierarchy reflects the updated intent rather than only updating text summaries in an
incorrect domain.

\paragraph{Algorithm.}
Algorithm~\ref{alg:modification} formalizes the differential modification procedure, including the branching logic
between in-place update and re-routing.

% -------------------------
% A.2.4 Insertion
% -------------------------
\subsubsection{Node Insertion via Semantic Routing}
\label{app:evolution_insert}

\input{algos/insertion}

\paragraph{Motivation.}
Naively attaching new entities to the root (or a fixed default module) breaks the semantic organization of the RPG and
degrades downstream navigation. Instead, we treat insertion as a \textbf{semantic placement} problem: find the most
appropriate abstract parent node in the current feature hierarchy that best matches the new entity’s functionality.

\paragraph{Routing objective.}
Let $u$ be a newly added code entity with semantic features $f(u)$. Starting from the root of the feature hierarchy,
we iteratively select the child domain whose description best aligns with $f(u)$, drilling down until no more
meaningful specialization is possible. This \textbf{top-down semantic routing} ensures that $u$ is inserted into the
most specific functional domain available while preserving interpretability of the hierarchy.

\paragraph{Algorithm.}
Algorithm~\ref{alg:insertion} formalizes the insertion procedure. At each step, the router considers the candidate
children of the current node and chooses the best target; if no child is sufficiently compatible, the algorithm
terminates and inserts $u$ at the current level. This prevents over-forcing entities into unrelated subtrees.

\paragraph{Complexity and scalability.}
Incremental evolution in RPG-Encoder is inherently local.
Each atomic operation affects only a bounded region of the hierarchy, without requiring global reconstruction.
As a result, maintenance cost scales with the \emph{magnitude of the change} rather than the size of the repository, enabling efficient and stable synchronization of the RPG under continuous development.

%% file: algos/deletion.tex
\begin{algorithm}[h]
\caption{Incremental Deletion (Recursive Pruning)}
\label{alg:deletion}
\begin{algorithmic}[1]
\Require Current graph $G$, target node id $id$
\Ensure Updated graph $G$

\Function{DeleteNode}{$G, id$}
    \State $v \gets \Call{GetNode}{G, id}$
    \If{$v = \bot$}
        \State \Return $G$
    \EndIf

    \State $parent \gets v.\text{parent}$
    \State \Call{RemoveNode}{G, v} \Comment{Remove node and incident edges}
    \State \Call{PruneOrphans}{G, parent} \Comment{Structural hygiene}
    \State \Return $G$
\EndFunction

\Function{PruneOrphans}{$G, v$}
    \If{$v = \bot \ \lor\  \Call{IsRoot}{v}$}
        \State \Return
    \EndIf

    \If{$\Call{IsEmpty}{v.\text{children}}$}
        \State $gp \gets v.\text{parent}$
        \State \Call{RemoveNode}{G, v} \Comment{Prune empty abstract category}
        \State \Call{PruneOrphans}{G, gp} \Comment{Recurse upwards}
    \EndIf
\EndFunction
\end{algorithmic}
\end{algorithm}

%% file: algos/modified.tex
\begin{algorithm}[htbp]
\caption{Differential Modification Handling}
\label{alg:modification}
\begin{algorithmic}[1]
\Require Graph $G$, file $f$, diff $\Delta$
\Ensure Updated graph $G$

\Function{ProcessModification}{$G, f, \Delta$}
    \State $\langle \mathcal{U}^+, \mathcal{U}^-, \mathcal{U}^{\sim} \rangle \gets \Call{ParseUnitDiff}{\Delta}$

    \Statex \textbf{1) Delete / Insert}
    \ForAll{$u \in \mathcal{U}^-$}
        \State $G \gets \Call{DeleteNode}{G, u.\text{id}}$
    \EndFor
    \ForAll{$u \in \mathcal{U}^+$}
      \State $G \gets \Call{InsertNode}{G, u, \textsc{LLMExtract}(u)}$
    \EndFor

    \Statex \textbf{2) Update / Re-route}
    \ForAll{$u \in \mathcal{U}^{\sim}$}
        \State $v \gets \Call{GetNode}{G, u.\text{id}};\; v.f \gets \Call{LLMUpdate}{u}$
        \If{$\Call{SemanticShift}{v} > \tau_{\text{drift}}$}
            \State $G \gets \Call{DeleteNode}{G, u.\text{id}}$ \Comment{logic drift}
           \State $G \gets \Call{InsertNode}{G, u, \textsc{LLMExtract}(u)}$
        \EndIf
    \EndFor

    \State \Return $G$
\EndFunction
\end{algorithmic}
\end{algorithm}

%% file: algos/insertion.tex
\begin{algorithm}[h]
\caption{Incremental Additions (LLM-Based Semantic Routing)}
\label{alg:insertion}
\begin{algorithmic}[1]
\Require Current Graph $G$, New Unit $u$, Feature Summary $f_u$
\Ensure Updated Graph $G$ with $u$ inserted

\Function{InsertNode}{$G, u, f_u$}
    \State $v_{\text{best}} \gets \Call{FindBestParent}{G.\text{root}, f_u}$
    \State $v_{\text{new}} \gets \Call{CreateNode}{u, f_u}$
    \State \Call{AddEdge}{$G, v_{\text{best}}, v_{\text{new}}, \mathcal{E}_{\text{feature}}$}
        \Comment{Attach to semantically determined parent}
    \State \Return $G$
\EndFunction

\Statex
\Function{FindBestParent}{$v_{\text{curr}}, f_{\text{target}}$}
    \State $Candidates \gets \{\,c \in \Call{Children}{v_{\text{curr}}} \mid \Call{IsAbstract}{c}\,\}$
    
    \If{$Candidates = \emptyset$}
        \State \Return $v_{\text{curr}}$ \Comment{Base case: No deeper abstract categories}
    \EndIf

    \State \Comment{Prompt LLM to select the best functional fit among children}
    \State $Context \gets \{ (c.\text{id}, c.f) \mid c \in Candidates \}$
    \State $v_{\text{choice}} \gets \Call{LLM\_Route}{Context, f_{\text{target}}}$
    
    \If{$v_{\text{choice}} \neq \text{null}$}
        \State \Return $\Call{FindBestParent}{v_{\text{choice}}, f_{\text{target}}}$
            \Comment{LLM chose a branch, drill down recursively}
    \Else
        \State \Return $v_{\text{curr}}$ \Comment{LLM decided no child is a better fit}
    \EndIf
\EndFunction
\end{algorithmic}
\end{algorithm}

%% file: appendix/methods/operation.tex
% =========================
% A.3 RPG Operation
% =========================
\subsection{RPG Operation: Agentic Tool-use and Navigation Logic}
\label{app:operation}

This subsection details how RPG is operationalized as an actionable substrate for repository understanding.
Beyond serving as a semantic representation, RPG exposes a \emph{tool interface} that bridges high-level intents
to concrete code entities and their dependency contexts. Concretely, we provide three complementary tools:
\textbf{SearchNode} for intent-based discovery, \textbf{FetchNode} for precision context retrieval, and
\textbf{ExploreRPG} for structural traversal on the RPG topology.

% -------------------------
% A.3.1 Tool Interfaces
% -------------------------
\subsubsection{Tool Interfaces and Prompt Specifications}
\label{app:operation_interfaces}

\paragraph{Design principles.}
The tool suite is designed to support a common agent workflow in repository understanding:
(i) start from vague or behavioral intents and obtain candidate code anchors;
(ii) verify anchors with precise source context; and
(iii) expand locally to cover call chains and related components.
To ensure tool outputs are deterministic and machine-consumable, each tool prompt defines a strict parameter schema
and return format.

\paragraph{SearchNode: intent-based discovery.}
\textbf{SearchNode} unifies \emph{semantic discovery} and \emph{textual retrieval}. It supports three modes:
\texttt{features} (intent $\rightarrow$ feature nodes / mapped code entities),
\texttt{snippets} (keyword/symbol search over the repository),
and \texttt{auto} (feature mapping first, followed by snippet search when needed).
Importantly, \texttt{search\_scopes} can restrict the search to selected feature subtrees, leveraging the grounded
hierarchy constructed in Appendix~\ref{app:extraction} to improve precision.

% Option A: move your existing tcblisting into separate files and input here
\input{prompts/tool_searchnode}

\paragraph{FetchNode: precision retrieval and verification.}
\textbf{FetchNode} retrieves exact source context and metadata for known candidates (code entities or feature paths).
It is used as a verification step after discovery to ensure the agent reasons on faithful code snippets rather than
speculative guesses. FetchNode returns file paths, line ranges, entity types, mapped feature information, and a code preview.

\input{prompts/tool_fetchnode}

\paragraph{ExploreRPG: topological traversal.}
\textbf{ExploreRPG} exposes the structural connectivity of RPG, enabling traversal along dependency edges
(\texttt{imports}, \texttt{invokes}, \texttt{inherits}, etc.) and/or containment/composition relations.
Starting from validated anchors, the agent can traverse upstream/downstream to uncover dependencies, impacted components,
and semantically related regions.

\input{prompts/tool_explorerpg}

% -------------------------
% A.3.2 Tool-use Policy
% -------------------------
\subsubsection{Tool-use Policy for Repository Understanding}
\label{app:operation_policy}

\paragraph{Canonical tool orchestration.}
We adopt a simple and robust orchestration policy that prioritizes semantic grounding before reading large contexts.
Given a natural-language intent $\mathcal{I}$, the agent executes:

\begin{enumerate}
    \item \textbf{Semantic discovery (SearchNode/features or auto):}
    convert $\mathcal{I}$ into concrete behavioral terms and retrieve candidate feature nodes and mapped code entities.
    If available, supply \texttt{search\_scopes} to restrict discovery to the most relevant functional subtrees.

    \item \textbf{Precision verification (FetchNode):}
    for top candidates, fetch exact code context (file path + line range + preview) and confirm semantic compatibility.
    Candidates that cannot be verified are discarded.

    \item \textbf{Local expansion (ExploreRPG):}
    from verified anchors, traverse dependency edges (e.g., \texttt{invokes}, \texttt{imports}) to recover call chains,
    utilities, and related modules. This step is used to (i) locate the root cause, (ii) map the impact surface, or
    (iii) identify integration points.

    \item \textbf{Pinpoint retrieval (optional SearchNode/snippets):}
    if the target remains ambiguous, run snippet search with high-signal identifiers obtained from previous steps
    (exact symbols, file paths, error strings), optionally extracting specific line ranges.
\end{enumerate}

\paragraph{Fallback rules.}
When semantic discovery returns insufficient recall (e.g., missing/weak feature matches), the agent falls back to
\texttt{snippets} mode to bootstrap concrete anchors, then returns to \textbf{FetchNode} and \textbf{ExploreRPG}.
When snippet search yields too many matches, the agent tightens constraints by adding (i) feature scopes,
(ii) file path patterns, or (iii) symbol-qualified queries.

This policy minimizes wasted context and reduces hallucination risk:
SearchNode provides intent-to-code grounding, FetchNode ensures the agent reasons on exact source, and ExploreRPG
reveals topological structure that cannot be reliably inferred from local snippets alone.

% -------------------------
% A.3.3 Execution Traces / Examples
% -------------------------
\subsubsection{Execution Traces and Examples}
\label{app:operation_examples}

\begin{figure}[htbp]
    \centering
    \includegraphics[width=\linewidth]{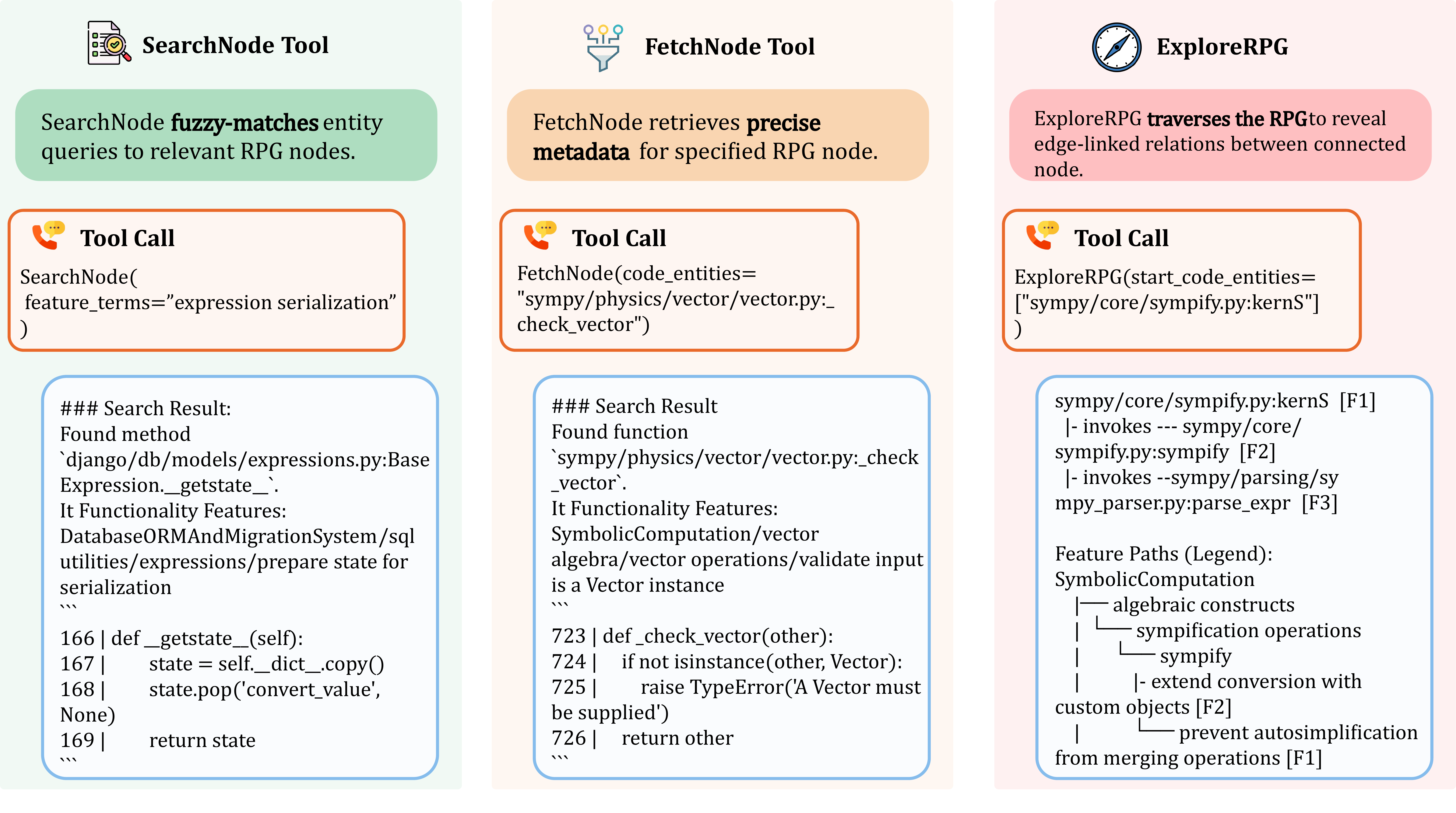}
    \caption{Execution traces of the three primary agentic tools. \textbf{SearchNode} maps abstract intent to concrete code;
    \textbf{FetchNode} retrieves precise source context; and \textbf{ExploreRPG} reveals topological connections and call relations.}
    \label{fig:tool_execution_examples}
\end{figure}

We illustrate the practical efficacy of these tools through the execution traces shown in
Figure~\ref{fig:tool_execution_examples}. These traces demonstrate how the agent navigates from abstract intents to
specific code implementations, leveraging both the semantic hierarchy and the dependency topology of RPG.

As depicted in the figure, each tool provides distinct structural signals that support the agent's reasoning:

\begin{itemize}
    \item \textbf{SearchNode (Left):}
    demonstrates intent-to-code grounding by mapping a behavioral query (e.g., ``expression serialization'') to a
    concrete code entity and its associated feature description. This step transforms ambiguous intent into executable anchors.

    \item \textbf{FetchNode (Center):}
    retrieves precise source context for a candidate entity (e.g., \texttt{\_check\_vector}), including exact line ranges
    and a preview snippet, enabling verification and preventing reasoning on speculative locations.

    \item \textbf{ExploreRPG (Right):}
    traverses the RPG topology from a verified anchor (e.g., \texttt{kernS}) to expose invocation and dependency relations.
    By showing edges such as \texttt{invokes} and their connected nodes, the agent can recover call chains and impacted modules,
    supporting systematic debugging and repository-level understanding.
\end{itemize}

%% file: prompts/tool_searchnode.tex
\begin{tcblisting}{
  title={SearchNode Tool Prompt},
  colback=lightgray,
  colframe=black,
  arc=1mm,
  boxrule=1pt,
  left=1mm,right=1mm,top=1mm,bottom=1mm,
  breakable,
  fontupper=\scriptsize\ttfamily,
  listing only,
  listing engine=listings,
  listing options={
    breaklines=true,
    breakatwhitespace=false,
    breakindent=0pt,
    prebreak=\mbox{},
    postbreak=\mbox{},
    keepspaces=true,
    columns=fullflexible,
    tabsize=4
  }
}
## Tool Name: SearchNode
### Description
Unified search tool for repository navigation. Use it to (1) map high-level functional/behavioral descriptions to concrete code entities via RPG mapping, and/or (2) retrieve concrete code snippets via symbol/file/keyword search. Prefer behavior-to-code mapping when you don't know the exact file/class/function name; then narrow down with snippet search.
Tip: Avoid vague terms; use concrete behavior phrases or high-signal identifiers.
### Parameters
{
  "tool_name": "SearchNode",
  "parameters": {
    "mode": "<'features' | 'snippets' | 'auto'. Required. 'auto' may run both: feature-mapping first, then snippet search.>",
    "feature_terms": "<List of concrete behavioral/functionality phrases. Required when mode is 'features' or 'auto'.>",
    "search_scopes": "<List of valid feature entity paths to restrict the Functionality SubGraph. Optional.>",
    "search_terms": "<List of file paths, qualified entities (file:Class.method), or high-signal text keywords. Required when mode is 'snippets' or when 'auto' proceeds to snippet search.>",
    "line_nums": "<Two integers [start, end] to extract lines from a specific file. Requires an exact file path. Optional.>",
    "file_path_or_pattern": "<File path or glob pattern to restrict snippet search. Default: '**/*.py'>",
  }
}
### Returns
- If feature search runs: matched feature nodes mapped to code entities (feature name, code entity, file path, line range when available)
- If snippet search runs: matched code snippets, complete files, or located entities based on search terms / line ranges
\end{tcblisting}

%% file: prompts/tool_fetchnode.tex
\begin{tcblisting}{
  title={FetchNode Tool Prompt},
  colback=lightgray,
  colframe=black,
  arc=1mm,
  boxrule=1pt,
  left=1mm,right=1mm,top=1mm,bottom=1mm,
  breakable,
  fontupper=\scriptsize\ttfamily,
  listing only,
  listing engine=listings,
  listing options={
    breaklines=true,
    breakatwhitespace=false,
    breakindent=0pt,
    prebreak=\mbox{},
    postbreak=\mbox{},
    keepspaces=true,
    columns=fullflexible,
    tabsize=4
  }
}
## Tool Name: FetchNode
### Description
- Retrieve precise metadata and source context for known code or feature entities.
- Use this tool to verify candidate code locations after identifying them through searches or graph exploration.
- Returns exact file path, entity type, start/end lines, mapped feature information, and a code preview.

### Parameters
{
  "tool_name": "FetchNode",
  "parameters": {
    "code_entities": "<List of existing and validated code entities in the current repository; non-existent paths or speculative entities may be ignored. Optional.>",
    "feature_entities": "<List of existing and validated feature paths in the current repository; non-existent entries may be ignored. Optional.>"
  }
}
### Returns
- Entity type (file/class/method/feature)
- Feature paths and code content (with source context / preview)
- Start/end lines and mapped feature information (when available)
\end{tcblisting}

%% file: prompts/tool_explorerpg.tex
\begin{tcblisting}{
  title={ExploreRPG Tool Prompt},
  colback=lightgray,
  colframe=black,
  arc=1mm,
  boxrule=1pt,
  left=1mm,right=1mm,top=1mm,bottom=1mm,
  breakable,
  fontupper=\scriptsize\ttfamily,
  listing only,
  listing engine=listings,
  listing options={
    breaklines=true,
    breakatwhitespace=false,
    breakindent=0pt,
    prebreak=\mbox{},
    postbreak=\mbox{},
    keepspaces=true,
    columns=fullflexible,
    tabsize=4
  }
}
## Tool Name: ExploreRPG
### Description
- Explore call chains and functional paths in the Repository Planning Graph.
- Starting from known code or feature entities, traverse upstream/downstream to discover related functions, files, and feature nodes.
### Parameters
{
  "tool_name": "ExploreRPG",
  "parameters": {
    "start_code_entities": "<Optional list of existing code entities in the current repository (file paths, classes, functions, or qualified names). Non-existent/speculative entities may be ignored or rejected.>",
    "start_feature_entities": "<Optional list of existing feature paths in the current repository. Non-existent entries may be ignored or rejected.>",
    "direction": "<Traversal direction: 'upstream' (dependencies), 'downstream' (dependents), or 'both'. Default: 'downstream'.>",
    "traversal_depth": "<Maximum traversal depth. Default: 2. Use -1 for unlimited depth.>",
    "entity_type_filter": "<Optional filter restricting traversal node types. Valid values: 'directory', 'file', 'class', 'function', 'method'.>",
    "dependency_type_filter": "<Optional filter restricting dependency edge types. Valid values: 'composes', 'contains', 'inherits', 'invokes', 'imports'.>"
  }
}
### Returns
- Connected nodes and edges (code or feature view)
- Hints for invalid or fuzzy matches
\end{tcblisting}

%% file: appendix/exp_setup.tex
\section{Experiment Setup}
\label{app:exp_setup}

This appendix provides additional experimental setup details. It is organized into two parts:
(i) repository reconstruction and (ii) repository understanding, including detailed baseline configurations and formal metric definitions.

% ============================================================
\subsection{Repository Understanding}
\label{app:under_detail}

\subsubsection{Experiment Setup}
\label{app:under_params_baselines}

We describe the implementation details and baseline configurations for the repository understanding task.
Our goal is to facilitate reproducibility and ensure fair and controlled comparisons across different localization pipelines.

\paragraph{Common evaluation protocol.}
All methods are evaluated under a shared protocol, including identical datasets, evaluation metrics, and termination criteria.
Unless otherwise specified, we use the same preprocessing, canonicalization, and ranking-based evaluation procedures described in Section~\ref{app:under_metrics}.

\paragraph{Backbone models.}
We evaluate multiple large language model backbones to assess the robustness of each localization pipeline to the underlying model choice, including
\emph{o3-mini}(\texttt{o3-mini-20250131})~\citep{o3-mini}, \emph{GPT-4o}(gpt-4o-20241120)~\citep{gpt4o-system-card}, \emph{GPT-4.1}(gpt-4.1-20250414)~\citep{gpt4.1}, \emph{GPT-5}(gpt-5-20250807)~\citep{gpt5-system-card},
\emph{DeepSeek-V3.1}~\citep{deepseek-v3.1}, and \emph{Claude-Sonnet-4.5}~\citep{anthropic_claude_sonnet_4_5}.

\paragraph{Baselines.}
We compare against representative repository-level localization pipelines:
\textbf{Agentless}~\citep{xia2024agentless},
\textbf{LocAgent}~\citep{chen2025locagent},
\textbf{CoSIL}~\citep{jiang2025cosil},
and \textbf{OrcaLoca}~\citep{yu2025orcaloca}.
For each baseline, we retain the original algorithmic structure and design choices, making only minimal and necessary adaptations to the benchmark interface and backbone model to ensure compatibility with the shared evaluation protocol.

% ----------------------------
\paragraph{Agentless} Agentless~\citep{xia2024agentless} employs a staged non-agentic workflow: (1) \textbf{Direct Prediction}: LLM predicts suspicious files directly from the issue description. (2) \textbf{Filtered Retrieval}: It performs embedding search within a search space pruned of "irrelevant folders." (3) \textbf{Candidate Aggregation}: Results from both streams are merged to maximize file-level recall.(4) \textbf{Element Localization}: Granularity is narrowed from files to specific code elements. (5) \textbf{Edit Localization}: The system pinpoints line-level edit targets within those elements. To ensure reproducibility, we apply specific parameter constraints corresponding to these stages. \textbf{Globally}, across all ranking steps, we maintain \texttt{top\_n=10} and enforce determinism via \texttt{num\_samples=1}. \textbf{Stage-specific configurations} are set as follows: For \textbf{Retrieval (Step 2)}, we employ \texttt{jinaai/jina-embeddings-v3} as the embedding backbone and set \texttt{filter\_type="given\_files"} to strictly enforce the LLM-generated folder constraints. For \textbf{Fine-grained Localization (Steps 4-5)}, we enable the \texttt{--compress} flag, which optimizes context utilization by condensing verbose code details while preserving salient information for precise element and edit identification.

% ----------------------------
\paragraph{LocAgent} LocAgent~\citep{chen2025locagent} is a dependency-graph integrated agent framework for repository-level localization, which wraps the dependency graph into three tools: (1) \textbf{SearchEntity}: searches relevant files/classes/functions from text queries (supports fuzzy match). (2) \textbf{TraverseGraph}: multi-hop traverses dependency relations from a seed entity to surface connected candidates. (3) \textbf{RetrieveEntity}: fetches the full metadata and code of selected entities for final inspection and ranking. For LocAgent, we do not impose any restriction on the number of iterative search rounds. To maximize the chance of producing a valid final prediction, we set the maximum retry budget to $3$ attempts, and take the first well-formed output that satisfies the evaluation interface. We run LocAgent in \textbf{function-calling} mode with parallelism set to $1$.
We impose no explicit limit on the number of iterative search steps, and set the maximum retry budget to $3$ to maximize the chance of producing a valid final output.
% TODO: Describe graph-guided traversal and ranking.
% TODO: Graph construction: dependency graph? call graph? file graph?
% TODO: Budget mapping: traversal steps / ranking iterations under T.

% ----------------------------
\paragraph{CoSIL} CoSIL~\citep{jiang2025cosil} adopts a hybrid \emph{agentic and workflow} strategy that explores code dependencies via iterative call-graph searching: it first performs broad exploration with a module call graph, then expands to a function call graph for deeper search, while using pruning and reflection to control direction and stabilize tool-formatted outputs. Following CoSIL’s implementation details, We run CoSIL in \textbf{function-calling} mode with parallelism set to $1$. We do not explicitly cap its iterative graph-search rounds, and allow up to $3$ retries to maximize the chance of obtaining a valid final output.

% TODO: Describe graph-based search / localization.
% TODO: Graph definition and query mechanism.
% TODO: Budget mapping under T.

% ----------------------------
\paragraph{OrcaLoca} OrcaLoca~\citep{yu2025orcaloca} combines agentic code-graph exploration with a \textbf{dynamic-analysis} signal, using \emph{bug reproduction} and \emph{regression tests} to guide iterative search and candidate verification. It introduces two key mechanisms: (1) \textbf{Action decomposition} factorizes the large search action space into a hierarchical decision process (e.g., first selecting candidate classes, then narrowing to files), and applies top-$k$ selection for class decomposition and file decomposition; (2) \textbf{Distance-aware context pruning} retains only a fixed budget of the most relevant context entries ($12$ in our setup), prioritizing code units that are closer to the current targets in the dependency/call graph to improve context efficiency. We follow the original setup: for action decomposition, it applies top-$k$ selection with $k{=}3$ for class decomposition and $k{=}2$ for file decomposition, and uses distance-aware context pruning with a budget of $12$ retained entries.

% TODO: Describe exploration and planning loop (browse/search/read/decide).
% TODO: Tooling (file open, search) and how each action is counted as a step.
% TODO: Budget mapping under T.

% TODO: Provide concrete settings used in experiments

% ============================================================
\subsubsection{Evaluation Targets at Multiple Granularities}
To assess localization quality at different levels of abstraction, we evaluate predictions at two granularities using a unified canonicalization scheme.

Each predicted or ground-truth location is mapped to a canonical string key through a granularity-specific formatter, ensuring consistent comparison across methods.

\paragraph{File-level.}
At the file level, a location is represented by its relative file path, \emph{e.g.},
\texttt{path/to/file.py}. For an instance $i$, both the ground-truth set
$G^{\text{file}}_i$ and the ranked prediction list $\pi^{\text{file}}_i$
consist of file paths. File-level evaluation measures whether a method can correctly identify the source files that contain the relevant implementation.

\paragraph{Function-level.}
At the function level, a location is represented by a fully qualified entity identifier within a file, formatted as \texttt{file:entity}. To avoid artificial mismatches caused by syntactic variations, constructor annotations are normalized by removing the suffix
\texttt{.\_\_init\_\_} when present. For example,
\texttt{a/b.py:Foo.\_\_init\_\_} is canonicalized to \texttt{a/b.py:Foo}.

Function-level evaluation assesses whether a method can precisely localize the relevant function or class definition beyond the file boundary.

When function-level annotations are unavailable for a given instance, we restrict the evaluation to the file level.

For both ground-truth and predicted locations, we remove duplicate entries while preserving their original order before computing all ranking metrics.

% ============================================================
\subsubsection{Metrics}
\label{app:under_metrics}

We formalize the ranking-based evaluation protocol for file-level localization.
Let $\mathcal{I}$ denote the set of evaluation instances. For each instance
$i \in \mathcal{I}$:
\begin{itemize}
    \item $G_i$ denotes the set of ground-truth relevant files (or locations),
    with cardinality $|G_i| = m_i$.
    \item $\pi_i = (p_{i,1}, p_{i,2}, \ldots, p_{i,|\pi_i|})$ denotes the ranked
    prediction list produced by a method.
\end{itemize}

We define a binary hit indicator sequence $\mathbf{h}_i \in \{0,1\}^{|\pi_i|}$ as
\begin{equation}
    h_{i,j} =
    \begin{cases}
        1, & \text{if } p_{i,j} \in G_i,\\
        0, & \text{otherwise},
    \end{cases}
    \quad j = 1,\ldots,|\pi_i|.
\end{equation}

All metrics are computed per instance and then averaged over $\mathcal{I}$.

\paragraph{Accuracy@k (Acc@k).}
Accuracy@k measures whether at least one ground-truth item appears within the top-$k$ predictions:
\begin{equation}
\mathrm{Acc@k} =
\frac{1}{|\mathcal{I}|}
\sum_{i\in\mathcal{I}}
\mathbb{I}\!\left[
\sum_{j=1}^{k} h_{i,j} \ge 1
\right].
\end{equation}
In our experiments, we report results for $k \in \{1, 3, 5\}$.

\paragraph{Recall.}
Recall measures the fraction of ground-truth items that are successfully retrieved by the model across the entire ranked list:
\begin{equation}
\mathrm{Recall} =
\frac{1}{|\mathcal{I}|}
\sum_{i\in\mathcal{I}}
\begin{cases}
\frac{\sum_{j=1}^{|\pi_i|} h_{i,j}}{|G_i|}, & |G_i| > 0,\\
0, & |G_i| = 0.
\end{cases}
\end{equation}

\paragraph{Precision.}
Precision measures the proportion of correct predictions among all retrieved items:
\begin{equation}
\mathrm{Precision} =
\frac{1}{|\mathcal{I}|}
\sum_{i\in\mathcal{I}}
\frac{\sum_{j=1}^{|\pi_i|} h_{i,j}}{|\pi_i|}.
\end{equation}

% ============================================================
\subsection{Details about Repository Reconstruction}
\label{app:recon_detail}
In this section, we provide a comprehensive description of the experimental setup, workflow logic, and termination protocols for the two comparative settings: ZeroRepo-Doc and ZeroRepo-RPG.

\subsubsection{RepoCraft Benchmark Construction}
To rigorous evaluate the capabilities of automated repository reconstruction, we adapted the \textbf{RepoCraft} benchmark. The benchmark consists of real-world Python repositories selected for their popularity and structural complexity. 

\begin{figure}[ht]
    \centering
    \includegraphics[width=1.0\linewidth]{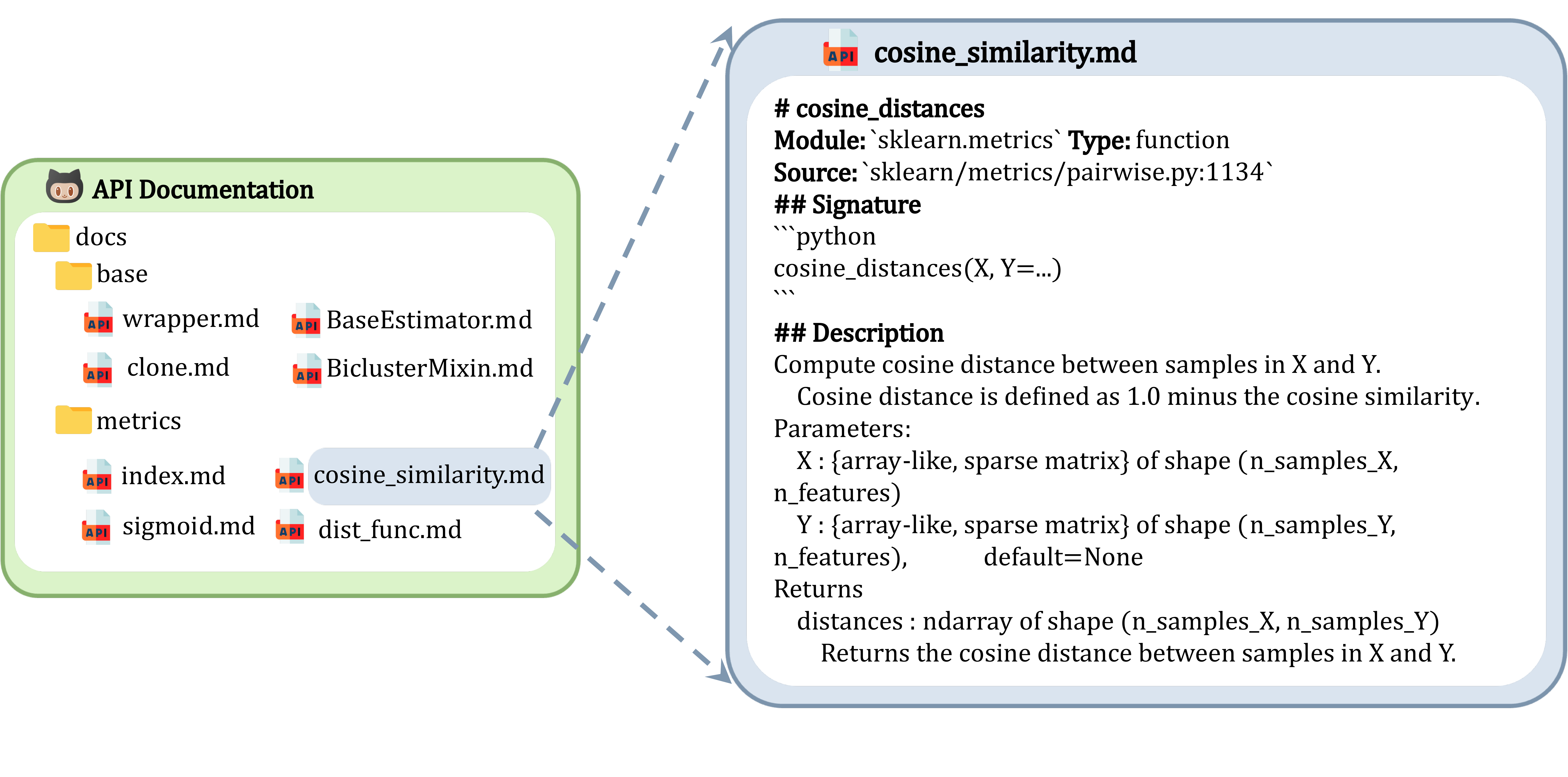}
    \caption{Illustration of the hierarchical organization of the API documentation (left) and an example of the granular content within a specific documentation node (right), detailing function signatures and parameter descriptions.}
    \label{fig:doc_structure}
\end{figure}

\input{tables/api_doc_stat}

\paragraph{Documentation Compilation.}
A critical component of our control setting is the provision of high-quality, official API documentation to serve as the ground-truth specification. We constructed this documentation dataset by processing the source files located in the \texttt{docs/} directory of each target repository. Specifically, we utilized \textbf{Sphinx}, the standard Python documentation generator, to compile the raw reStructuredText (reST) or Markdown files into a unified textual representation, as illustrated in Figure~\ref{fig:doc_structure}. This compiled documentation captures the official definitions of classes, functions, and module hierarchies, ensuring that the baseline agents operate on the exact same informational standard as human developers reading the official manual. As summarized in Table~\ref{tab:repo_stats}, the scale of this context is substantial: the compiled documentation spans a total of 7,320 files and over 2.5 million tokens across the six subject repositories, presenting a rigorous benchmark for unstructured long-context understanding.

\subsubsection{Baselines}
\label{app:recon_baseline}

\begin{figure*}[t]
    \centering
    \includegraphics[width=\linewidth]{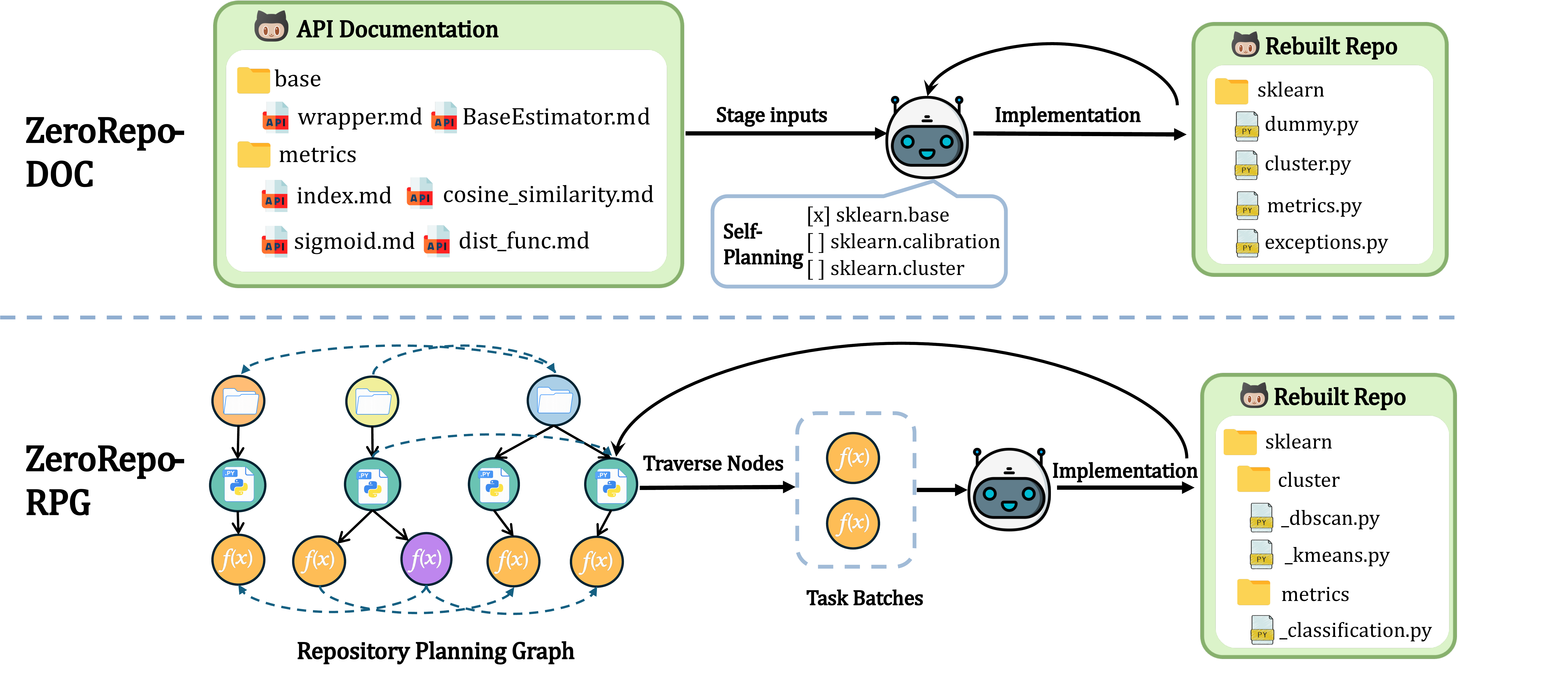} % 请替换为你的实际文件名
    \caption{Comparison of reconstruction workflows adapted from ZeroRepo. \textbf{Top (ZeroRepo-Doc):} The baseline relies on unstructured API documentation, requiring the agent to perform autonomous self-planning and manual state tracking. \textbf{Bottom (ZeroRepo-RPG):} Our method leverages the extracted RPG as a structured prior, enabling deterministic topological traversal and context-aware batched execution.}
    \label{fig:workflow}
\end{figure*}

\paragraph{Backbone Framework: ZeroRepo.}
We adopt ZeroRepo \citep{luo2025rpg_zerorepo}, a state-of-the-art framework originally architected for zero-shot repository generation, as our unified execution backbone. In its native configuration, ZeroRepo operates by synthesizing a Repository Planning Graph (RPG) from high-level user intents. This graph serves as a rigid execution schedule, guiding the agent through a topological traversal where functionalities are implemented and verified via Test-Driven Development (TDD). The framework's modular design decouples \textit{planning} (Graph Construction) from \textit{execution} (Code Generation), making it an ideal substrate for isolating and comparing different planning strategies.

\paragraph{Adaptation for Reconstruction.}
For the task of Repository Reconstruction, the objective shifts from hallucinating new systems to recovering existing ground-truth architectures. To evaluate the efficacy of our extracted RPG against unstructured information, we adapt ZeroRepo into two distinct configurations by modifying its planning source:

\paragraph{ZeroRepo-Doc (Unstructured Baseline).}
In this configuration, we lobotomize the graph-based planning engine to simulate a conventional developer workflow. The agent is initialized solely with the raw API documentation of the target repository and an empty progress tracking file. Without a pre-computed graph, the reconstruction proceeds in an open-ended, iterative loop:
\begin{itemize}
    \item \textbf{Self-Planning}: In each iteration, the agent must manually cross-reference its progress against the documentation to identify pending modules and formulate its own immediate objectives.
    \item \textbf{Manual State Tracking}: The agent bears the full cognitive burden of global state maintenance, creating and updating a checklist to prevent redundant or missed implementations.
    \item \textbf{Execution}: It employs standard TDD, writing reproduction scripts and code based on its self-determined plan. The process terminates only when the agent subjectively judges that all documented requirements are met.
\end{itemize}

\paragraph{ZeroRepo-RPG (Graph-Guided Reconstruction).}
In our proposed configuration, we replace the generative planner with the \textbf{RPG-Encoder}. Instead of generating a graph from scratch, we inject the pre-extracted RPG as a ground-truth topological prior. This shifts the burden of planning from the agent to the substrate, transforming reconstruction into a deterministic traversal:
\begin{itemize}
    \item \textbf{Topological Traversal}: The extracted nodes are arranged into a strict \textbf{dependency-based topological order}. This guarantees that prerequisite modules are implemented before dependent ones, eliminating circular dependency errors.
    \item \textbf{LLM-Driven Batching}: To optimize efficiency, an LLM scheduler previews the topological queue and dynamically aggregates semantically coherent nodes (e.g., a set of related utility functions) into a single implementation batch.
    \item \textbf{Context-Aware Execution}: The system executes these batches sequentially. The agent is relieved of global planning and focuses purely on implementation, leveraging the explicit context ($f, \mathbf{m}$) provided by the graph nodes to generate code that aligns with the established architecture.
\end{itemize}

\subsubsection{Metrics}
To rigorously evaluate the fidelity of repository reconstruction, we adapt metrics from RepoCraft~\citep{luo2025rpg_zerorepo} to capture four complementary dimensions: \emph{functional coverage}, \emph{spectral novelty}, \emph{executable fidelity}, and \emph{structural scale}.
These metrics collectively assess (i) whether the ground-truth functionalities are successfully recovered, (ii) whether the model introduces extraneous capabilities, (iii) whether the reconstructed logic is semantically correct, and (iv) the complexity level of the realized system.

\paragraph{Functionality Coverage (Recovery Rate).} 
This metric quantifies the \textbf{recall} of the reconstruction process—specifically, what fraction of the ground-truth functional topology has been successfully restored.
We define the reference feature set $\mathcal{C} = \{c_1, \dots, c_K\}$ using the ground-truth repository's functional signature (derived from developer documentation or canonical feature lists).
The generated functionalities $\mathcal{G} = \{g_1, \dots, g_N\}$ are extracted from the reconstructed codebase.
To measure alignment, we employ K-Means clustering with $\mathcal{C}$ as fixed centroids, mapping each generated feature $g_i$ to its nearest reference intent $f(g_i) \in \mathcal{C} \cup \{c_{\text{OOD}}\}$ (where $c_{\text{OOD}}$ denotes Out-Of-Distribution).
Feature assignments are further refined by an LLM-as-Judge to mitigate semantic drift.
Coverage is defined as the proportion of ground-truth categories "hit" by the reconstruction:
\begin{equation}
\text{Coverage} = \frac{1}{|\mathcal{C}|} \sum_{j=1}^{K} \mathbbm{1}\left[\exists g_i \in \mathcal{G}, \ f(g_i) = c_j \right].
\end{equation}
A higher coverage indicates that the RPG successfully guided the agent to rebuild a more complete functional footprint of the original system.

\paragraph{Functionality Novelty (Extrapolation).} 
In the context of reconstruction, pure imitation is not always possible; models may "hallucinate" valid but unrequested features.
To capture this behavior, we measure the proportion of generated functionalities that fall outside the ground-truth taxonomy.
Novelty is calculated as the fraction of generated nodes assigned to the $c_{\text{OOD}}$ centroid:
\begin{equation}
\text{Novelty} = \frac{1}{|\mathcal{G}|} \sum_{i=1}^{N} \mathbbm{1}\left[f(g_i) = c_{\text{OOD}} \right].
\end{equation}
While high novelty is desirable in open-ended generation, for reconstruction tasks, it characterizes the model's tendency to extrapolate or deviate from the strict blueprint provided by the RPG.

\paragraph{Functionality Accuracy (Executable Fidelity).} 
Coverage validates intent, but fidelity requires correctness.
We evaluate the semantic equivalence of the reconstructed code against the original repository's behavior using adapted test suites.
We report two statistics:
\begin{itemize}
    \item \textbf{Voting Rate (Feature Presence)}: The fraction of tasks where the validation pipeline confirms that a plausible implementation of the target algorithm exists. This measures the system's ability to \emph{instantiate} the planned logic.
    \item \textbf{Success Rate (Semantic Correctness)}: The fraction of tasks where the reconstructed code passes the unit tests. This measures the \emph{executable validity} of the implementation.
\end{itemize}

\paragraph{Code-Level Statistics (Structural Scale).} 
Finally, to ensure the reconstruction is not merely a skeletal prototype, we compare the scale of the generated codebase against realistic standards.
Metrics are computed over filtered source files (excluding non-production artifacts like \texttt{tests} or \texttt{benchmarks}) to assess structural complexity:
\begin{itemize}
    \item \textbf{File Count}: Reflects the modular granularity and file-level organization.
    \item \textbf{Normalized LOC}: Effective lines of code (excluding comments/blanks), proxying implementation volume.
    \item \textbf{Code Token Count}: Total lexical tokens, indicating the density and complexity of the synthesized logic.
\end{itemize}

\subsubsection{Model Configuration} 
\label{app:recon_impl}

To strictly separate structural reasoning from coding capability, we utilize distinct models for the planning and execution phases. We employ \textbf{GPT-4o}~\citep{gpt4o-system-card} as the extraction engine to analyze the source repository and generate the Repository Planning Graph (RPG), leveraging its strong reasoning abilities to ensure high-fidelity structural representation. For the downstream reconstruction tasks within the ZeroRepo framework, we evaluate two different backbone models: \textbf{GPT-5-mini} and \textbf{GPT-4.1}. This variation allows us to assess the robustness of our method across models with differing parameter scales.

%% file: tables/api_doc_stat.tex
\begin{table}[ht]
\centering
\caption{Statistics of the compiled API documentation across the six subject repositories.}
\label{tab:repo_stats}
% \small 
\renewcommand{\arraystretch}{1.1} 
\setlength{\tabcolsep}{5pt}
\begin{tabular}{lcccccc}
\toprule
\textbf{Metric} & \textbf{Django} & \textbf{Pandas} & \textbf{Requests} & \textbf{Sklearn} & \textbf{Stats} & \textbf{SymPy} \\
\midrule
\textbf{\# Files} & 2,863 & 1,536 & 15 & 1,052 & 1,244 & 610 \\
\textbf{\# Tokens} & 374,586 & 1,314,314 & 52,351 & 235,319 & 405,022 & 180,505 \\
\bottomrule
\end{tabular}
\end{table}

%% file: appendix/results.tex
\section{More Results}
\label{app:results}

\subsection{Repository Reconstruction}
\label{app:results_recon}
In this section, we provide detailed per-repository results on RepoCraft, reporting performance for each repository individually to complement the main aggregated statistics.
\input{tables/full_reconstruction}

\paragraph{Analysis of Repository-Specific Performance.} The breakdown of results across six diverse repositories reveals consistent patterns confirming the scalability of \ours{}. First, in large-scale scientific libraries like Scikit-Learn and Sympy, the documentation-based baseline suffers from severe structural collapse, often recovering less than 20\% of the file structure (e.g., only 29 files for Sympy with GPT-4.1). In contrast, \ours{} leverages hierarchical planning to maintain architectural integrity, reconstructing code volumes that closely approximate the human-written gold standard (e.g., generating $\sim$900k tokens for Scikit-Learn). Second, for frameworks with intricate inter-dependencies like Django, \ours{} achieves 100\% coverage and a pass rate exceeding 96\%, demonstrating that topological constraints effectively guide the agent through complex logic flows where linear documentation fails. Ultimately, across all datasets, \ours{} consistently bridges the gap between sparse intent and dense implementation, proving its robustness as a domain-agnostic substrate for repository reconstruction.

\subsection{Agent Behavior}
\label{app:results_behavior}

\begin{figure}[ht]
    \centering
    \includegraphics[width=1.0\linewidth]{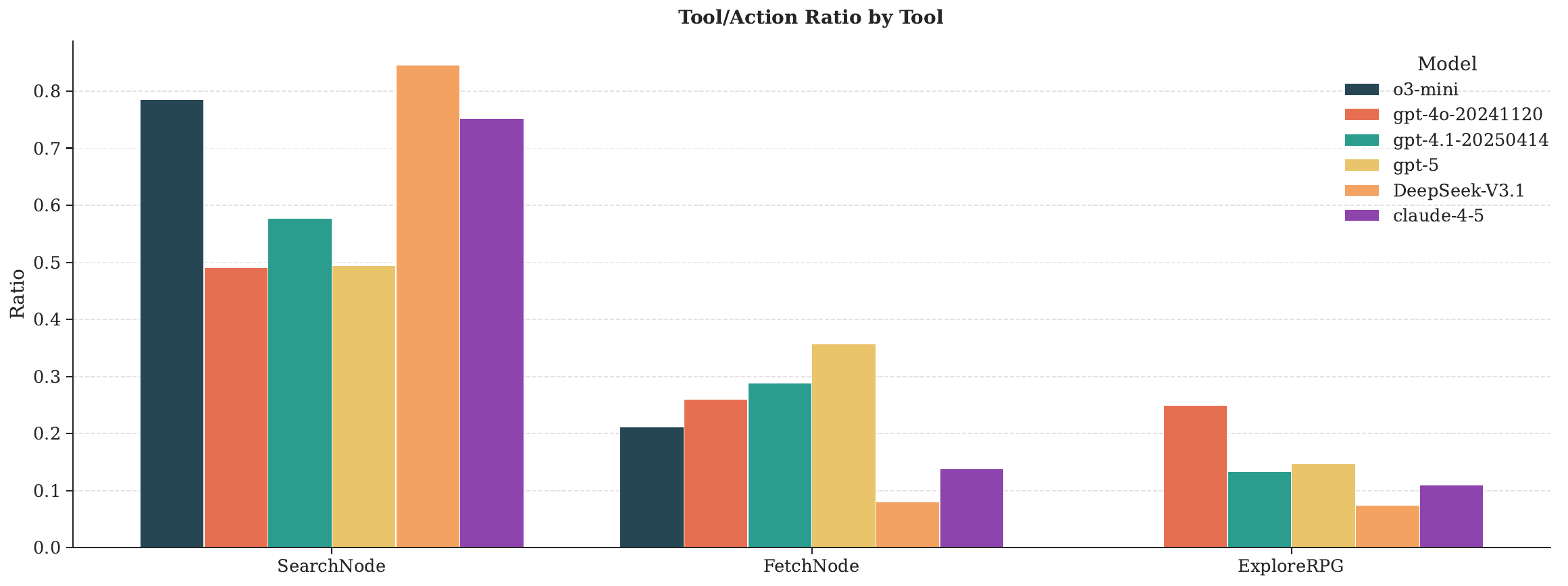} 
    \caption{\label{fig:tool_usage_analysis} Breakdown of the Tool/Action ratio for each tool across different models. This metric reflects the frequency of specific tool invocations relative to the total actions performed.}
\end{figure}

We calculated the tool-to-action ratios across all evaluation episodes to quantify the behavioral strategies of each model, as shown in Figure \ref{fig:tool_usage_analysis}. The results indicate that the agents generally maintain a robust balance between keyword-based retrieval and structural navigation. While \texttt{SearchNode} serves as a foundational tool for locating initial entry points, the significant utilization of \texttt{ExploreRPG}—especially by high-performing models like GPT-4o—highlights its critical role in the reasoning loop. This distribution suggests that capable agents effectively recognize the value of \texttt{ExploreRPG}, strategically employing it to leverage topological connections and gain a holistic understanding of the codebase, rather than relying solely on local keyword matches.

\subsection{Error Analysis}
\label{app:err_analysis}

\input{tables/error_analysis}

To look beyond aggregate metrics and understand the behavioral divergence between RPG-enhanced agents and baselines, we conducted a manual qualitative analysis on a stratified sample of failure cases. Based on the taxonomy defined in Table~\ref{tab:error_taxonomy}, we observe distinct error patterns that highlight the structural advantages of RPG while exposing persistent challenges in agentic reasoning.

\begin{figure}[htbp]
    \centering
    % 请确保文件名与你项目中的实际文件名一致
    \includegraphics[width=\linewidth]{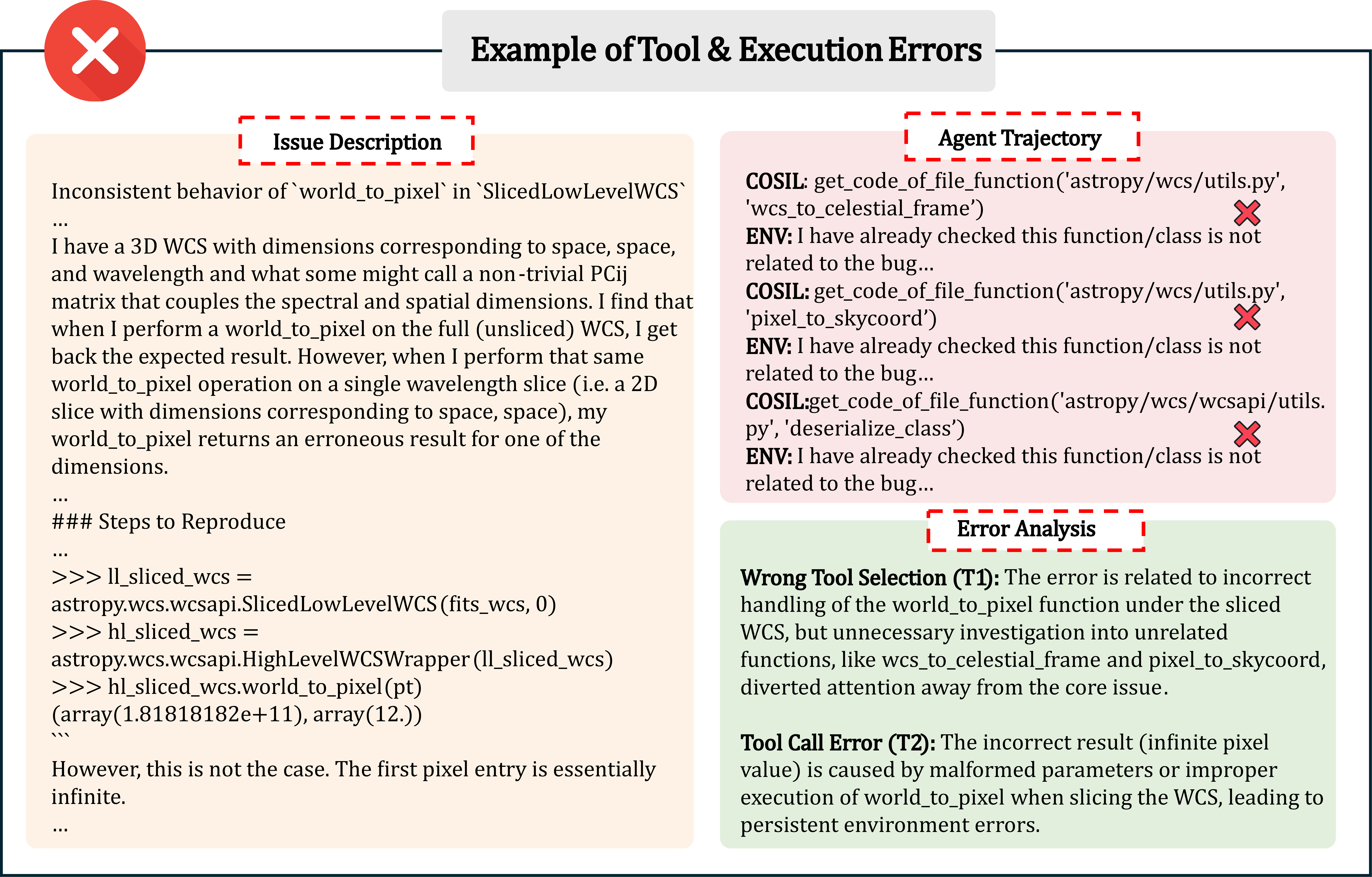}
    \caption{Case study of tool misuse and execution failure. COSIL exhibits \textbf{Wrong Tool Selection} by investigating unrelated functions (e.g., \texttt{wcs\_to\_celestial\_frame}) instead of the core issue (`world\_to\_pixel`), leading to persistent environment errors and a failure to reproduce the bug.}
    \label{fig:t1_error_case_study}
    \vspace*{-10pt}
\end{figure}

\begin{figure*}[htbp]
    \centering
    \includegraphics[width=\textwidth]{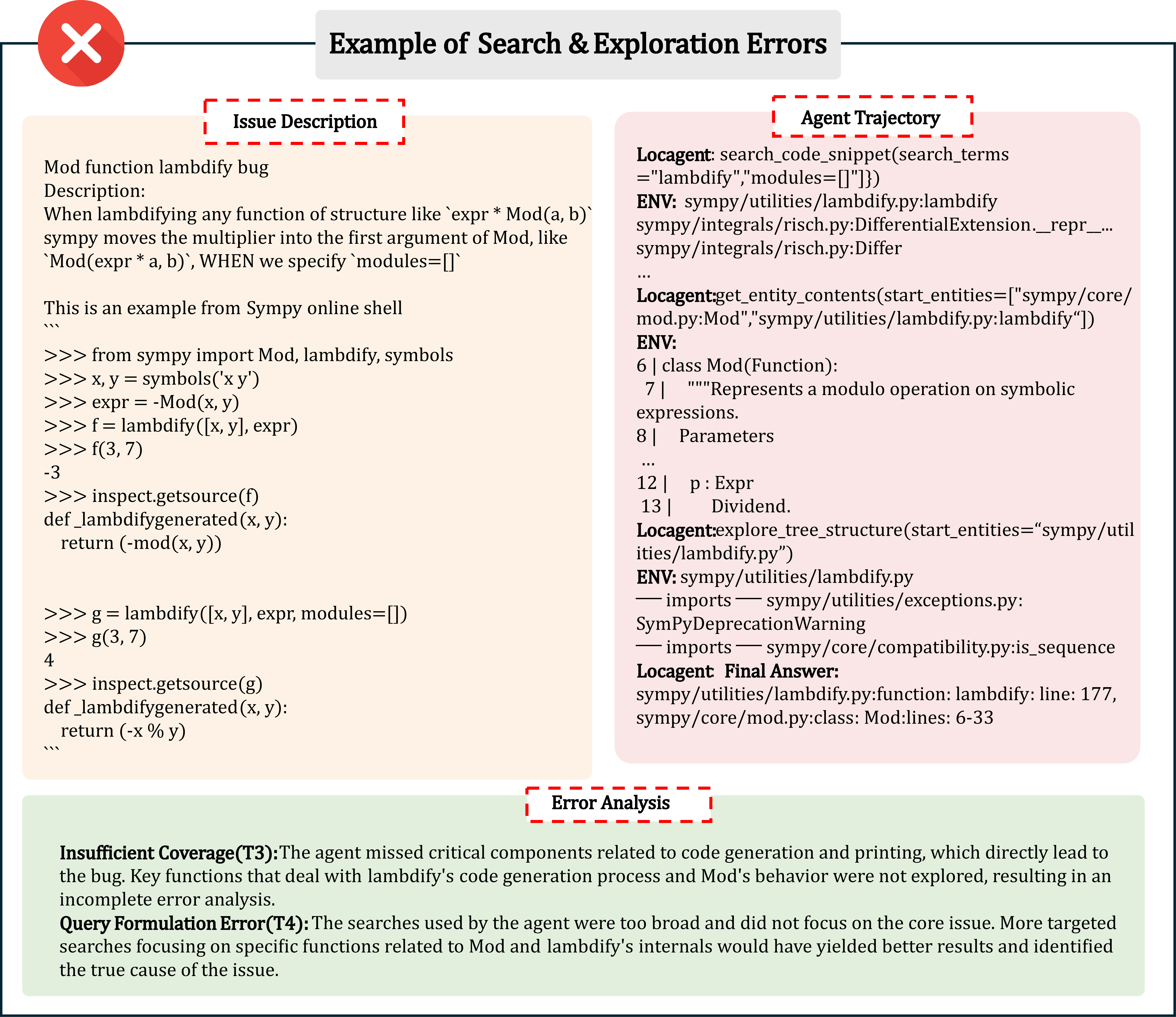}
    \caption{Case study of a search and exploration failure. The agent attempts to resolve a \texttt{lambdify} bug in SymPy but fails due to \textbf{Insufficient Coverage (T3)}, missing critical code generation logic, and \textbf{Query Formulation Errors (T4)}, where searches were too broad to locate the root cause.}
    \label{fig:t3_error_case_study}
    \vspace*{-10pt}
\end{figure*}

\subsection{Cost Analysis}
\label{app:cost_analysis}

\input{tables/cost_full}

Beyond effectiveness and behavioral patterns, we further analyze the computational cost of different agents to assess their practical efficiency. We report the average token consumption and monetary cost per instance across all evaluation episodes.

Overall, RPG-enhanced agents incur moderately higher costs than lightweight baselines due to more frequent structured exploration and reasoning steps. However, this increase is well-controlled and scales sublinearly with performance gains, indicating that improved task success is not achieved through excessive tool usage or redundant interactions. In particular, high-performing models demonstrate a favorable cost--performance trade-off by allocating additional budget primarily to informative exploration rather than repeated failed actions.

We note that for \textsc{LocAgent} with \texttt{o3-mini}, the reported cost can be unusually low, as the agent often terminates after only 1--2 rounds and may occasionally produce an answer without invoking any tool. This reflects early termination behavior rather than efficient full-loop reasoning, and is specific to the \textsc{LocAgent}--\texttt{o3-mini} pairing.

For completeness, we provide the full cost breakdown for all models, including prompt tokens, completion tokens, and estimated monetary cost, in Table~\ref{tab:full_efficiency_analysis}.

%% file: tables/full_reconstruction.tex
\FloatBarrier

% =================Requests Repo=================
\begin{table}[ht]
\centering
\caption{Main results on \textbf{Requests} reconstruction tasks in RepoCraft.}
\label{tab:requests_full_reconstruction}
\small
\renewcommand{\arraystretch}{0.88}
\setlength{\tabcolsep}{4pt}      
\resizebox{\textwidth}{!}{%
\begin{tabular}{llccccc}
\toprule
\textbf{Framework} & \textbf{Backbone} & \textbf{Coverage} (\%) $\uparrow$ & \textbf{Pass Rate} (\%) $\uparrow$ & \textbf{\#Files} $\uparrow$ & \textbf{nLOC} $\uparrow$ & \textbf{Code Tokens} $\uparrow$ \\
\midrule
\rowcolor{modelbg}
\multirow{1}{*}{Gold Projects (Reference)} 
  & Human Developers 
  & 100.0
  & 93.0 / 93.0
  & 17
  & 2273
  & 22,297 \\
\midrule
\multirow{2}{*}{ZeroRepo-Doc (Baseline)}
  & GPT-4.1    & 54.6 & 13.2 / 32.1 & 7 & 968 & 8141 \\
  & GPT-5-mini & 72.7 & 27.5 / 54.9 & 28 & 1,890 & 15,842 \\
\midrule
\multirow{2}{*}{\textbf{ZeroRepo-RPG (Ours)}}
  & GPT-4.1    & 100.0 & 88.4 / 95.3 & 23 & 2901 & 24,734 \\
  & GPT-5-mini & 100.0 & 88.4 / 95.3 & 20 & 3503 & 27,188 \\
\bottomrule
\end{tabular}%
}
\vspace*{-5pt}
\end{table}

% =================Sklearn Repo=================
\begin{table}[ht]
\centering
\caption{Main results on \textbf{Scikit-Learn} reconstruction tasks in RepoCraft.}
\label{tab:sklearn_full_reconstruction}
\small
\renewcommand{\arraystretch}{0.88}
\setlength{\tabcolsep}{4pt}        
\resizebox{\textwidth}{!}{%
\begin{tabular}{llccccc}
\toprule
\textbf{Framework} & \textbf{Backbone} & \textbf{Coverage} (\%) $\uparrow$ & \textbf{Pass Rate} (\%) $\uparrow$ & \textbf{\#Files} $\uparrow$ & \textbf{nLOC} $\uparrow$ & \textbf{Code Tokens} $\uparrow$ \\
\midrule
\rowcolor{modelbg}
\multirow{1}{*}{Gold Projects (Reference)} 
  & Human Developers 
  & 100.0
  & 99.5 / 100.0
  & 185
  & 65,927
  & 592,187 \\
\midrule
\multirow{2}{*}{ZeroRepo-Doc (Baseline)}
  & GPT-4.1    & 68.1 & 53.9 / 64.0 & 181 & 6,787 & 69,468 \\
  & GPT-5-mini & 72.3 & 55.6 / 66.3 & 76 & 12,007 & 101,988 \\
\midrule
\multirow{2}{*}{\textbf{ZeroRepo-RPG (Ours)}}
  & GPT-4.1    & 100.0 &  79.3 / 99.5 & 269 & 71,835 & 698,602 \\
  & GPT-5-mini & 100.0 &  82.8 / 99.5 & 267  & 96,806  & 900,381  \\
\bottomrule
\end{tabular}%
}
\vspace*{-5pt}
\end{table}

% =================Sympy Repo=================
\begin{table}[ht]
\centering
\caption{Main results on \textbf{Sympy} reconstruction tasks in RepoCraft.}
\label{tab:sympy_full_reconstruction}
\small
\renewcommand{\arraystretch}{0.88} 
\setlength{\tabcolsep}{4pt} 
\resizebox{\textwidth}{!}{%
\begin{tabular}{llccccc}
\toprule
\textbf{Framework} & \textbf{Backbone} & \textbf{Coverage} (\%) $\uparrow$ & \textbf{Pass Rate} (\%) $\uparrow$ & \textbf{\#Files} $\uparrow$ & \textbf{nLOC} $\uparrow$ & \textbf{Code Tokens} $\uparrow$ \\
\midrule
\rowcolor{modelbg}
\multirow{1}{*}{Gold Projects (Reference)} 
  & Human Developers 
  & 100.0
  & 97.8 / 100.0
  & 699
  & 218,924
  & 943,873 \\
\midrule
\multirow{2}{*}{ZeroRepo-Doc (Baseline)}
  & GPT-4.1    & 33.3 & 36.4 / 48.2 & 29 & 605 & 5347 \\
  & GPT-5-mini & 66.7 & 52.5 / 64.7 & 135 & 10,454 & 99,460 \\
\midrule
\multirow{2}{*}{\textbf{ZeroRepo-RPG (Ours)}}
  & GPT-4.1    & 91.7 & 63.6 / 89.6 & 220 & 59,283 & 565,680 \\
  & GPT-5-mini & 95.8 & 81.4 / 96.6 & 288 & 80,534 & 686,951  \\
\bottomrule
\end{tabular}%
}
\vspace*{-5pt}
\end{table}

% =================Statsmodels Repo=================
\begin{table}[ht]
\centering
\caption{Main results on \textbf{Statsmodels} reconstruction tasks in RepoCraft.}
\label{tab:statsmodels_full_reconstruction}
\small
\renewcommand{\arraystretch}{0.88}
\setlength{\tabcolsep}{4pt}        
\resizebox{\textwidth}{!}{%
\begin{tabular}{llccccc}
\toprule
\textbf{Framework} & \textbf{Backbone} & \textbf{Coverage} (\%) $\uparrow$ & \textbf{Pass Rate} (\%) $\uparrow$ & \textbf{\#Files} $\uparrow$ & \textbf{nLOC} $\uparrow$ & \textbf{Code Tokens} $\uparrow$ \\
\midrule
\rowcolor{modelbg}
\multirow{1}{*}{Gold Projects (Reference)} 
  & Human Developers 
  & 100.0
  & 96.8 / 100.0
  & 271
  & 83,325
  & 893,824 \\
\midrule
\multirow{2}{*}{ZeroRepo-Doc (Baseline)}
  & GPT-4.1    & 66.7 & 71.7 / 75.9 & 87 & 5,235 & 55,688 \\
  & GPT-5-mini & 75.0 & 57.8 / 79.5 & 163 & 25,472 & 274,983 \\
\midrule
\multirow{2}{*}{\textbf{ZeroRepo-RPG (Ours)}}
  & GPT-4.1    & 96.6 & 89.8 / 98.9 & 303 & 66,045 & 709,424 \\
  & GPT-5-mini & 97.7 & 91.4 / 100.0 & 297 & 92,618 & 942,411 \\
\bottomrule
\end{tabular}%
}
\vspace*{-5pt}
\end{table}

% =================Pandas Repo=================
\begin{table}[ht]
\centering
\caption{Main results on \textbf{Pandas} reconstruction tasks in RepoCraft.}
\label{tab:pandas_full_reconstruction}
\small
\renewcommand{\arraystretch}{0.88}
\setlength{\tabcolsep}{4pt}    
\resizebox{\textwidth}{!}{%
\begin{tabular}{llccccc}
\toprule
\textbf{Framework} & \textbf{Backbone} & \textbf{Coverage} (\%) $\uparrow$ & \textbf{Pass Rate} (\%) $\uparrow$ & \textbf{\#Files} $\uparrow$ & \textbf{nLOC} $\uparrow$ & \textbf{Code Tokens} $\uparrow$ \\
\midrule
\rowcolor{modelbg}
\multirow{1}{*}{Gold Projects (Reference)} 
  & Human Developers 
  & 100.0
  & 99.0 / 100.0
  & 217
  & 106,447
  & 943,873 \\
\midrule
\multirow{2}{*}{ZeroRepo-Doc (Baseline)}
  & GPT-4.1    & 81.8 & 63.4 / 82.5 & 726 & 14,770 & 127,326 \\
  & GPT-5-mini & 75.0 & 56.2 / 81.4 & 180 & 14,610 & 125,957 \\
\midrule
\multirow{2}{*}{\textbf{ZeroRepo-RPG (Ours)}}
  & GPT-4.1    & 97.4 & 89.5 / 92.8 & 146 & 21,657 & 201,484 \\
  & GPT-5-mini & 97.4 & 80.1 / 97.4 & 115 & 30,066 & 249,094 \\
\bottomrule
\end{tabular}%
}
\vspace*{-5pt}
\end{table}

% =================Django Repo=================
\begin{table}[ht]
\centering
\caption{Main results on \textbf{Django} reconstruction tasks in RepoCraft.}
\label{tab:django_full_reconstruction}
\small
\renewcommand{\arraystretch}{0.88}
\setlength{\tabcolsep}{4pt}       
\resizebox{\textwidth}{!}{%
\begin{tabular}{llccccc}
\toprule
\textbf{Framework} & \textbf{Backbone} & \textbf{Coverage} (\%) $\uparrow$ & \textbf{Pass Rate} (\%) $\uparrow$ & \textbf{\#Files} $\uparrow$ & \textbf{nLOC} $\uparrow$ & \textbf{Code Tokens} $\uparrow$ \\
\midrule
\rowcolor{modelbg}
\multirow{1}{*}{Gold Projects (Reference)} 
  & Human Developers 
  & 100.0 
  & 83.0 / 100.0
  & 681
  & 109,457
  & 917,622 \\
\midrule
\multirow{2}{*}{ZeroRepo-Doc (Baseline)}
  & GPT-4.1    & 83.3 & 61.3 / 77.5 & 228 & 8114 & 687,718 \\
  & GPT-5-mini & 83.3 & 66.2 / 81.8 & 276 & 16,053 & 135,523 \\
\midrule
\multirow{2}{*}{\textbf{ZeroRepo-RPG (Ours)}}
  & GPT-4.1    & 100.0 & 96.2 / 98.1 & 408 & 43,468  & 391,260 \\
  & GPT-5-mini & 100.0 & 91.9 / 97.5 & 369 & 61,702  & 496,570 \\
\bottomrule
\end{tabular}%
}
\vspace*{-5pt}
\end{table}

\FloatBarrier

%% file: tables/error_analysis.tex
\begin{table*}[ht]
\centering
\caption{Error Taxonomy for Code Localization Agents (T1--T12)}
\label{tab:error_taxonomy}
\small
\renewcommand{\arraystretch}{1.3}
\newcolumntype{L}[1]{>{\RaggedRight\arraybackslash\hspace{0pt}}p{#1}}
\newcolumntype{Y}{>{\RaggedRight\arraybackslash\hspace{0pt}}X}

\begin{tabularx}{\textwidth}{@{} L{3.5cm} L{4.5cm} Y @{}}
\toprule
\textbf{Category} & \textbf{Error Type} & \textbf{Description} \\
\midrule

Tool \& Execution Errors
& Wrong Tool Selection (T1)
& Agent uses an inappropriate tool for the task; e.g., using \texttt{grep} when \texttt{find} is needed, or reading files when search is more suitable. \\
& Tool Call Error (T2)
& Malformed parameters, invalid paths/regex, or runtime failures (e.g., ``No such file'', syntax errors, timeouts). \\
\midrule

Search \& Exploration Errors
& Insufficient Coverage (T3)
& Agent fails to systematically explore relevant code areas; misses obvious entry points from the problem statement. \\
& Query Formulation Error (T4)
& Searches use wrong keywords, overly broad/narrow scope, or miss obvious terms from the problem statement. \\
& Redundant Search (T5)
& Agent makes repetitive tool calls; stuck in loops; repeats similar queries without progress. \\
\midrule

Reasoning \& Interpretation Errors
& Evidence Misinterpretation (T6)
& Agent misreads tool output; confuses similar symbols; incorrect understanding of code relationships. \\
& Info Synthesis Failure (T7)
& Agent fails to connect evidence across multiple steps; doesn't build a coherent picture from findings. \\
& Hallucination (T8)
& Agent invents non-existent files, functions, or paths; outputs fabricated information not grounded in evidence. \\
\midrule

Context \& Scope Errors
& Problem Misunderstanding (T9)
& Agent fundamentally misunderstands the task; focuses on the wrong aspect of the problem. \\
& Granularity Mismatch (T10)
& Output at the wrong specificity level (too coarse/fine); e.g., file-level when function-level is needed. \\
& Codebase Context Failure (T11)
& Agent fails to understand project structure, conventions, or architecture. \\
\midrule

Multi-Factor Errors
& Cascading Errors (T12)
& Multiple interacting errors where early mistakes propagate; recovery attempts fail. \\
\bottomrule
\end{tabularx}
\end{table*}

%% file: tables/cost_full.tex
\begin{table}[ht]
\centering
\caption{Efficiency analysis comparing the average number of agent interaction rounds and cost per instance.}
\label{tab:full_efficiency_analysis}
\small
\renewcommand{\arraystretch}{1.15}
\setlength{\tabcolsep}{0pt}

\begin{tabular*}{\columnwidth}{@{\extracolsep{\fill}} l l c c c }
\toprule
\textbf{LLM} & \textbf{Method} & \textbf{Steps} & \textbf{Cost (\$)} & \textbf{Eff.} \\
\midrule

% ================= o3-mini =================
\multirow{4}{*}{o3-mini}
& OrcaLoca           & 10.13 & 0.10 & 4.86 \\
& LocAgent           & \textbf{2.14}  & \textbf{0.04} & \textbf{21.75} \\
& CoSIL              & 18.95 & 0.18 & 4.07 \\
& \textbf{\ours{}}   & 6.53  & 0.11 & 7.07 \\
\cmidrule{1-5}

% ================= GPT-4o =================
\multirow{4}{*}{GPT-4o}
& OrcaLoca           & 29.26 & 0.98 & 0.54 \\
& LocAgent           & 8.95  & 0.50 & 1.35 \\
& CoSIL              & 20.65 & 0.30 & 2.21 \\
& \textbf{\ours{}}   & \textbf{8.22}  & \textbf{0.20} & \textbf{3.84} \\
\cmidrule{1-5}

% ================= GPT-4.1 =================
\multirow{4}{*}{GPT-4.1}
& OrcaLoca           & 20.22 & 0.46 & 1.48 \\
& LocAgent           & 11.94 & 0.86 & 0.76 \\
& CoSIL              & 19.77 & 0.24 & 3.10 \\
& \textbf{\ours{}}   & \textbf{6.75}  & \textbf{0.18} & \textbf{4.63} \\
\cmidrule{1-5}

% ================= GPT-5 =================
\multirow{4}{*}{GPT-5}
& OrcaLoca           & 36.93 & 0.75 & 1.16 \\
& LocAgent           & 6.48  & 0.49 & 1.64 \\
& CoSIL              & 19.52 & 0.31 & 2.64 \\
& \textbf{\ours{}}   & \textbf{6.34}  & \textbf{0.22} & \textbf{4.25} \\
\cmidrule{1-5}

% ================= Claude-4.5 =================
\multirow{4}{*}{Claude-4.5}
& OrcaLoca           & 11.32 & 0.88 & 0.91 \\
& LocAgent           & \textbf{7.54}  & 0.88 & 0.85 \\
& CoSIL              & 12.60 & \textbf{0.48} & \textbf{1.64} \\
& \textbf{\ours{}}   & 16.75 & 1.32 & 0.71 \\

\bottomrule
\end{tabular*}
\end{table}

%% file: appendix/ablations.tex
\section{Ablation}
\label{app:ablations}

\subsection{Repository Reconstruction}
\label{app:abl_recon}

\paragraph{Ablation Setup}
To rigorously isolate the contribution of hierarchical topological signals to repository reconstruction, we conduct a controlled ablation study on the \texttt{scikit-learn} repository from the \textbf{RepoCraft} benchmark. All experiments utilize \textbf{GPT-5-mini} as the underlying reasoning engine. We define two specific ablation settings to evaluate the agent's capability in structural inference:
\begin{itemize}
    \item \textbf{Function-Level Ablation (-Func):} In this setting, we strip fine-grained implementation guidance by removing metadata from leaf nodes (functions) and eliminating function-to-function dependency edges from the RPG. Consequently, the agent retains file-level boundaries but must autonomously deduce necessary function signatures from high-level features and independently derive their logical implementation order.
    \item \textbf{File-Level Ablation (-File/-Func):} Building upon the function-ablated graph, we further excise all file-level structural information, including file nodes and directory paths. This setting retains only the abstract semantic feature descriptions. It forces the agent to perform \textit{ab initio} structural organization: the agent must semantically aggregate discrete features to synthesize file architectures, design class and function hierarchies, and plan the global implementation sequence without any reference directory topology.
\end{itemize}

\begin{table}[htbp]
\centering
\caption{Feature Distribution across Different Modes}
\label{tab:feature_distribution}
\small
\setlength{\tabcolsep}{4.5pt}  
\renewcommand{\arraystretch}{1.15}

\begin{tabular}{l c c c c c c c c c}
\toprule
\multirow{2}{*}{\textbf{Category}} &
\multicolumn{3}{c}{\textbf{Full Mode}} &
\multicolumn{3}{c}{\textbf{Func Ablation}} &
\multicolumn{3}{c}{\textbf{File Ablation}} \\
\cmidrule(lr){2-4}\cmidrule(lr){5-7}\cmidrule(lr){8-10}
& \textbf{Count} & \textbf{Features} & \textbf{Avg$\pm$Std}
& \textbf{Count} & \textbf{Features} & \textbf{Avg$\pm$Std}
& \textbf{Count} & \textbf{Features} & \textbf{Avg$\pm$Std} \\
\midrule
File     & 250  & 4943 & $19.77\pm20.94$ & 275  & 6664 & $24.23\pm27.11$ & 154  & 5175 & $33.60\pm36.99$ \\
Class    & 524  & 3584 & $6.84\pm6.51$   & 815  & 3676 & $4.51\pm4.76$   & 466  & 2462 & $5.28\pm6.16$   \\
Function & 1366 & 1359 & $0.99\pm0.07$   & 1597 & 2988 & $1.87\pm2.07$   & 1202 & 2713 & $2.26\pm2.25$   \\
\bottomrule
\end{tabular}
\end{table}

\paragraph{Analysis of Structural Clustering Behaviors.}
Table~\ref{tab:feature_distribution} illustrates how different topological priors influence the LLM's organization strategies.
(1) \textbf{Granular Encapsulation in Function Ablation.} When function-level metadata is removed, the model exhibits a preference for explicit object-oriented modeling. The significant increase in class count (524 $\to$ 815) and function definitions indicates that without specific procedural guidance, the agent tends to atomize logic into smaller, discrete class-based units for encapsulation.
(2) \textbf{Feature Consolidation in File Ablation.} Conversely, the absence of file-level constraints shifts the behavior towards coarse-grained aggregation. The model consolidates features into fewer physical files (250 $\to$ 154), resulting in a higher feature density per file ($19.77 \to 33.60$). This suggests that without directory boundaries to enforce separation, the model inherently clusters semantically related functionalities into larger, unified modules rather than distributing them across a file system hierarchy.

\subsection{Repository Understanding}
\label{app:abl_understanding}

\paragraph{Ablation Setup}
To quantify the individual contributions of structural connectivity and semantic annotation, we evaluate \textbf{ZeroRepo-RPG} under two degradation protocols:
\begin{itemize}
    \item \textbf{Dependency Graph Ablation (w/o Dependency):} In this variant, we sever all static dependency edges ($\mathcal{E}_{\text{dep}}$) from the RPG. This ablation strictly limits the \texttt{ExploreRPG} tool by occluding execution logic; the agent loses the ability to traverse call graphs or trace upstream/downstream relationships, forcing it to navigate solely based on physical file hierarchy.

    \item \textbf{Semantic Feature Ablation (w/o Feature):} Here, we strip high-level feature descriptions ($f$) and functional subordination edges ($\mathcal{E}_{\text{feature}}$). This degradation fundamentally impairs semantic navigability: the \texttt{SearchNode} tool regresses from intent-based retrieval to rigid keyword matching, while \texttt{ExploreRPG} ceases to display functional hierarchies. Furthermore, retrieved nodes lack refined summaries, compelling the agent to infer utility solely from raw identifiers.
\end{itemize}

\paragraph{Efficiency Degradation from Component Loss.}
Table~\ref{tab:ablation_steps_cost} quantifies the operational overhead introduced by removing structural and semantic priors. The full \textbf{\ours{}} model consistently achieves the minimal trajectory length and cost across both LLMs, validating the synergistic efficiency of the complete graph. (1) \textbf{Impact of Semantic Features:} Removing feature metadata (\textit{w/o Features}) incurs the most significant penalty in exploration steps (e.g., rising from 8.22 to 9.23 on GPT-4o). Without high-level summaries to guide intent-based retrieval, the agent is forced into a trial-and-error loop, repeatedly fetching raw code to verify relevance, which drastically prolongs the search process.
(2) \textbf{Impact of Dependency Structure:} Severing dependency edges (\textit{w/o Dependency}) primarily inflates token cost (e.g., $+\$0.09$ on GPT-4.1). Lacking explicit execution paths, the agent must manually traverse the file system and read broader contexts to deduce logical connections, leading to inefficient information consumption compared to the surgical navigation enabled by the full RPG.

\begin{table}[htbp]
\centering
\caption{Impact of Dependency and Feature Ablations on Steps and Cost.}
\label{tab:ablation_steps_cost}
\small
\setlength{\tabcolsep}{5.5pt}
\renewcommand{\arraystretch}{1.15}

\begin{tabular}{lcc|cc}
\toprule
\multirow{2}{*}{\textbf{Setting}} &
\multicolumn{2}{c|}{\textbf{GPT-4o}} &
\multicolumn{2}{c}{\textbf{GPT-4.1}} \\
\cmidrule(lr){2-3}\cmidrule(lr){4-5}
& \textbf{Avg Steps} & \textbf{Cost (\$)} & \textbf{Avg Steps} & \textbf{Cost (\$)} \\
\midrule
\ours{}               & \textbf{8.22} & \textbf{0.20} & \textbf{6.75} & \textbf{0.26} \\
w/o Dependency            & 8.53 & 0.27 & 7.31 & 0.35 \\
w/o Features              & 9.23 & 0.30 & 7.37 & 0.37 \\
\bottomrule
\end{tabular}
\end{table}